\documentclass[final,1p,times,twocolumn, authoryear]{elsarticle}

\makeatletter
\def\ps@pprintTitle{%
  \let\@oddhead\@empty
  \let\@evenhead\@empty
  \let\@oddfoot\@empty
  \let\@evenfoot\@oddfoot
}
\makeatother
\usepackage{latexsym}
\usepackage{amssymb}
\usepackage{amsmath}
\usepackage{amsthm}
\usepackage{booktabs}
\usepackage{enumitem}
\usepackage{graphicx}
\usepackage{caption}
\usepackage{hyperref}
\usepackage{color}
\usepackage{array}
\usepackage{tablefootnote}
\usepackage{multirow}
\usepackage{parnotes}
\usepackage{lscape}
\usepackage{tabularx} 
\usepackage{lineno}

\newcommand{\BibTeX}{B\kern-.05em{\sc i\kern-.025em b}\kern-.08em\TeX}
\newcolumntype{P}[1]{>{\centering\arraybackslash}p{#1}}
\newcommand\bittiny{\fontsize{7}{10}\selectfont}

\journal{}

\begin{document}


\begin{frontmatter}

\title{Cooperative Patrol Routing:\\ Optimizing Urban Crime Surveillance through Multi-Agent Reinforcement Learning}

\author[A]{Juan Palma-Borda} 
\author[A]{Eduardo Guzmán}
\author[A]{María-Victoria Belmonte} 

\affiliation[A]{
            organization={Dpto. de Lenguajes y Ciencias de la Computación. E.T.S. de Ingeniería Informática. Universidad de Málaga},
            city={Málaga},
            country={Spain}}


\begin{abstract}
Patrolling can be defined as the act of visiting locations of interest at regular intervals for either surveillance, control, protection, or monitoring purposes. The effective design of patrol strategies is a difficult and complex problem, especially in medium and large areas. The objective is to plan, in a coordinated manner, the optimal routes for a set of patrols in a given area, in order to achieve maximum coverage of the area, while also trying to minimize the number of patrols. In this paper, we propose a multi-agent reinforcement learning (MARL) model, based on a decentralized partially observable Markov decision process, to plan unpredictable patrol routes within an urban environment represented as an undirected graph. The model attempts to maximize a target function that characterizes the environment within a given time frame. Our model has been tested to optimize police patrol routes in three medium-sized districts of the city of Malaga. The aim was to maximize surveillance coverage of the most crime-prone areas, based on actual crime data in the city. To address this problem, several MARL algorithms have been studied, and among these the \textit{Value Decomposition Proximal Policy Optimization} (VDPPO) algorithm exhibited the best performance. We also introduce a novel metric, the \textit{coverage index}, for the evaluation of the coverage performance of the routes generated by our model. This metric is inspired by the \textit{predictive accuracy index} (PAI), which is commonly used in criminology to detect hotspots. Using this metric, we have evaluated the model under various scenarios in which the number of agents (or patrols), their starting positions, and the level of information they can observe in the environment have been modified. Results show that the coordinated routes generated by our model achieve a coverage of more than $90\%$ of the $3\%$ of graph nodes with the highest crime incidence, and $65\%$ for $20\%$ of these nodes; $3\%$ and $20\%$ represent the coverage standards for police resource allocation. The source code of our implementation is available in this public repository (\href{https://github.com/iacomlab/marl-patrol-routing}{https://github.com/iacomlab/marl-patrol-routing}). The data cannot be provided for confidentiality reasons, as they contain sensitive information.

\end{abstract}



\end{frontmatter}

\section{Introduction}

Patrolling can be defined as the act of visiting locations of interest at regular intervals for surveillance, control, protection, or monitoring purposes \citep{machado2002multi, guo2023balancing}. This persistent monitoring process is repetitive in nature and takes prolonged periods of time \citep{hari2020optimal}. It can be applied to solve a wide range of problems in virtual worlds, such as strategic games, or in the real world: smart cities, smart defense, monitoring an area with drones, patrolling water resources, identifying objects or people in dangerous situations that should be rescued by humans or robots, etc. \citep{machado2002multi, luis2022evolutionary, soliman2023ai}. This is a multi-agent problem, since by nature it is conveniently well suited to be shared in space and time by several agents, leading to what is often called \textit{multi-agent patrolling} \citep{othmani2017multi}. A patrolling strategy that schedules visits to different areas by agents, is essential for the effective execution of the patrolling task. However, the design of patrol strategies is a notoriously difficult and complex problem \citep{guo2023balancing}. 

Patrolling strategies can be decomposed into three subtasks or problems \citep{samanta2022literature}: environment clusterization or district design, resource allocation, and route design. Concerning district design, it focuses on setting different disjunct areas within the environment where patrolling takes place. This clusterization is often developed using expert knowledge in terms of number of crimes committed, population density, rapid incident response, and other relevant socio-economic factors. Resource allocation consists in the identification of the optimal quantity of resources necessary to cover each design area, including manpower, vehicles, and other equipment. Finally, route design requires planning the optimal route for each patrol, in terms of the selected area and the available resources.

Regarding the design of the route, different types of patrolling could be required depending on the characteristics of the domain \citep{machado2002multi}. In this sense, the objectives may vary if the aim is to minimize response times to certain emergencies, maximize the number of hotspots visited, or minimize the cost of the route in terms of resources or delay time between two passes through the same place \citep{machado2002multi}. In addition, many situations demand that the routes be unpredictable, so that an adversary capable of observing the patrolling behavior of the agents cannot accurately predict the future locations \citep{guo2023balancing}, making the route design much more complex. Finally, it is important to bear in mind that there may be several patrols operating simultaneously and that the real benefit will result from the combination of these routes rather than from each route separately. Routes should therefore be planned in a coordinated manner.

In this paper, we will focus on the problem of unpredictable route design optimization in urban environments, aiming to maximize a target function without knowing the routes' final destinations and within a specific time frame. To tackle this problem, we propose a  multi-agent reinforcement learning (MARL) model based on a decentralized partially observable Markov decision process. This problem is studied following a cooperative approach with a discrete space action. The agents are homogeneous, i.e., there are no additional agents other than the patrols in the model, so the goal is shared by all the agents in the model. The observability of the model can be set as a partial or total, depending on whether full information about the area to be monitored or only a portion of this information is provided to the patrol agents. Our model is designed for environments where agents may have partial information about the environment. In this particular situation, the only information available to agents is the position of their peers prior to choosing an action. For this reason, we introduce the concept of agents' \emph{line of sight} as a way to determine the part of the surrounding environment that is known to them.

The environment in which patrols perform their activity strongly conditions this type of problem. The most complex scenario is the continuous patrolling of the terrain since the mobility space is large \citep{machado2002multi}. We follow the commonly used approach of digitizing the real terrain through a grid that is later transformed into a graph that models all the possible paths. This approach, called \emph{skeletonization}, leads to an abstract representation of the environment that can be applied to different types of problems \citep{machado2002multi}.

To evaluate the suitability and performance of our model, we have applied it to the concrete problem of police patrolling in a real urban environment. In this case, we start with different urban areas between 1.8 km\textsuperscript{2} and 3 km\textsuperscript{2} and a limited number of police patrols. The allocation of resources by public administrations is a topic that is subject to constant review because there is a general shortage of resources, and their misuse directly impacts the lives of citizens. There is not always an adequate number of human resources available to carry out effective policing, which leads to having to prioritize policing strategies in some areas of the city to the detriment of others.

The contributions of this work can be summarized as follows:
\begin{itemize}
    \item A new MARL-based model for the optimized and coordinated design of time-bound routes in urban environments without predetermining the specific nodes that will be part of the routes.

    \item The model can be used to estimate the optimal number of patrols, helping to avoid over-patrolling in high-tension areas and reducing the waste of public resources.
    
    \item Unlike most of the existing proposals in the literature, the model has been applied to a real urban problem: the optimization of police patrol routes in the city of Málaga, maximizing the number of high-crime areas monitored. To the best of our knowledge, this is the first approach designed to maximize the number of crimes monitored in a coordinated manner, given a set of patrols during a single shift. 
    
    \item Most metrics evaluating route design are based on the concept of idleness, which is however meaningless in our domain. For this reason, to evaluate the performance of our model, we introduce a novel metric, the \textit{coverage index}, to rate the coverage performance of the routes. We have analyzed the model performance in terms of agents' number, observability, and starting position.
\end{itemize}

The paper is organized as follows: Section 2 describes the state of the art in the field of patrol route design, focusing on those approaches that use MARL solutions. Section 3 formalizes our approach to the problem as a decentralized partially observable Markov decision process. Section 4 outlines how our model can be applied to optimize urban crime surveillance in three urban environments, also highlighting our studies on the most appropriate MARL algorithm for this problem. Section 5 presents the results of the experiments in these urban areas, and Section 6 discusses them. Finally, Section 7 presents some conclusions and future work.

\section{Related Works}

In this section, we review the state of the art in the field of patrol routing design, focusing on those approaches that use MARL. Multi-agent solutions for patrolling areas have a well-established presence in the literature, but the introduction of new algorithms of MARL in these solutions is not widely considered due to their high training costs in terms of time and resources and their relatively recent development. In the first subsection, we introduce basic features of MARL and then go on to describe some of the most prominent works in the literature, most of which belong to the domain of multi-robot patrolling. 

\subsection{Multi-Agent Reinforcement Learning}

MARL stands as a relevant framework for acquiring the agent policies required to accomplish specific tasks across diverse domains, encompassing disaster response \citep{parker2016exploiting}, autonomous vehicles \citep{shalev2016safe}, or multiplayer games \citep{wijaya2019multiagent, samvelyan2019starcraft}. To complete this task, several algorithms can be found in the literature, principally grouped into 4 families: Proximal Policy Optimization (PPO) \citep{schulman2017proximal}, Deep Q-Network (DQN) \citep{mnih2015human} or Deep Deterministic Policy Gradient (DDPG) \citep{lillicrap2015continuous}, Advantage Actor Critic (A2C) \citep{mnih2016asynchronous}, and Trust Region Policy Optimization (TRPO) \citep{schulman2015trust}. 

The selection of the most suitable algorithm for a certain problem to be solved is generally conditioned by the limitations of each model. Some algorithms may not be applicable to continuous or discrete action spaces or may not be designed for competitive or for cooperative problems. For cooperative problems, the approaches and constraints of MARL algorithms are even more important, primarily due to the necessity of coordinating various agents to achieve a certain task or several tasks, and many of them are only achievable when agents perform various tasks in a coordinated manner. This coordination of agents is not trivial, especially in problems with homogeneous agents, where there are no associated roles, which would allow decomposing the problem into subtasks that agents in each role could learn. Finally, some recent approaches have introduced MARL algorithms for solving problems with incomplete information, as a way to train agents in particular environments such as natural disasters. These approaches perform well compared to usual solving techniques, especially if they are complemented with expert knowledge of the subject \citep{LEE2021296}. 

\subsection{Patrol Route Design}

The route design problem focuses on planning the path of one or more patrols, depending on the selected area of action and the available resources. This task aims to achieve maximum area coverage at minimal cost in a coordinated manner between assigned patrols, thus giving coordination an essential role. In addition, in some cases it may be necessary to introduce a certain degree of randomness \citet{yin2012trusts}.

One of the main areas of application of the patrol route design problem is robotics, more specifically multi-robot patrolling, which is perhaps the most explored field in terms of the study of routes within a certain area. In the literature, various techniques for planning these robots using centralized or distributed coordination can be found \citep{huang2019survey}. 
Centralized coordination includes those methods managed by a central coordinator \citep{machado2002multicoor,almeida2003combining,othmani2017multi} or cyclic strategies \citep{chevaleyre2004theoretical} where robots are arranged on a closed path. For instance, \citet{pasqualetti2012cooperative} used this technique to generate a route to visit all important locations. Partition-based strategy \citep{chevaleyre2004theoretical} is another centralized coordination strategy, where the area is divided into disjoint regions, similarly to the environment clusterization but in smaller regions. Using this strategy, each agent covers one region of the divided area, having as many regions as agents. \citet{portugal2010msp} establishes a \textit{multilevel subgraph patrolling} (MSP) algorithm by modifying the partitioning phase, thus achieving a reduction in redundant work compared to cyclic approach. On another note, \citet{sea2018frequency} uses k-means clustering to divide the regions and find the shortest path using a simulated annealing algorithm. Lastly, \citet{STRANDERS201363} presents several environments based on the elaboration of patrol routes on graphs using both multi- and single-agent approaches. This divides the entire graph into different groups called atomic clusters, for which a solution is sought that determines the travel time and establishes the entry and exit points of each cluster.

Distributed coordination includes three groups: 1) Reactive approaches \citep{machado2002multicoor}, where each agent selects the nodes with the most idleness (this metric quantifies the duration for which a node remains unvisited) aiming to reduce it. For example, \citet{yan2016multi} proposes a distributed algorithm where each robot estimates the idleness of each vertex using information shared with the other agents, without the need for centralized planning or control, and \citet{chen2015designing} develops a new police planning strategy that mixes Bayesian and ant colony methods and tries to minimize the average time lag between consecutive visits to hotpots. 2) Auction-based approaches implement negotiation mechanisms among the robots. For example, \citet{hwang2009cooperative} explores the coordination of patrols and develops a system where each patrol selects the points to visit through an auction system. 3) Learning-based approaches, where the agents adapt their strategies to the environment having only partial information \citep{santana2004multi, othmani2018, othmani2019, portugal2013applying, portugal2016cooperative, guo2023balancing}. 
\citet{santana2004multi} models this problem as a semi-Markov decision problem using reinforcement learning as a way to coordinate all agents behaviors; \citet{othmani2018, othmani2019} proposes an LSTM (\emph{long-short term memory}) architecture using the \textit{heuristic pathfinder cognitive coordinated} strategy combined with small randomness to train the agents;
 \citet{portugal2013applying} and \citet{portugal2016cooperative} propose a \textit{concurrent Bayesian learning strategy} (CBLS) and develop a probabilistic model to represent the availability of robots to move to a neighboring point from their current location. Additionally, they also adopt a reward-based technique to influence the robots' future movements during patrolling; \citet{guo2023balancing} have recently addressed another key factor, i.e., balancing unpredictability and efficiency along with a MARL approach, where the agents are trained using HAPPO (\textit{heterogeneous-agent proximal policy optimization}); lastly, \citet{chen2023risk} establishes a distributed model combined with the cyclic strategy often found in centralized approaches. In this model, agents select their routes based on the crime values assigned to the roads in question. 

Table \ref{table:literature_review} summarizes the main features of the aforementioned works, highlighting the differentiating characteristics of our approach, which are detailed in the last row of the table. The first column of this table shows the reference of each proposal; the second one summarizes the main purpose of the proposal; and the third shows the type of agent for which the proposal was developed, robot or human. The fourth column indicates whether the environment used is real or artificial. In this sense, if an environment used in the simulation is designated as an existing city center, neighborhood, institution, building, or vehicle, this environment is labeled as real. The model behind the proposal is described in the fifth column, and column six registers if coordination between agents is centralized or distributed. The seventh column classifies the size of the studied area. The term "Tiny" is used to describe a street or a floor of a building; "Small" to describe a campus, a housing estate, or a single building; "Medium" to describe an entire neighborhood or group of neighborhoods; and "Large" to characterize a medium-sized city or district of a metropolis. The last two columns indicate whether the nodes or points to be monitored are predetermined within the selected area, or whether each node is of equal importance to be surveilled, and the metrics used to evaluate the performance of the proposal.

Next, we outline the main differences between our approach and those presented in the table. These differences highlight how our works diverge in terms of assumptions and objectives, addressing challenges in urban patrolling for human patrols that have been overlooked or simplified in related studies, primarily due to the use of robots instead of humans as agents in the model.

\begin{itemize}
    \item Most of the approaches \citep{pasqualetti2012cooperative, pasqualetti2012cooperative2, STRANDERS201363, portugal2016cooperative, othmani2017multi, othmani2018, othmani2019} focus on reducing the idleness of all nodes in the environment or a preselection of them without prioritizing visits to more problematic areas or those of greater interest. For example, to achieve this goal, \citet{othmani2017multi, othmani2018, othmani2019} generates a multi-agent LSTM model on different types of graphs, while \citet{portugal2013applying} and \citet{portugal2016cooperative} applies Bayesian learning in order to achieve the same goal. In our work, nodes are not preselected, neither is it feasible to cover all of them within a single shift, making this metric inapplicable. In light of this, our objective is to identify the optimal route that ensures comprehensive coverage of the areas of most interest.
    
    \item Many works are designed for continuous monitoring of an area, regardless of work shifts or a finite time frame \citep{STRANDERS201363, yan2016multi}. In our approach, human patrols are carried out in 8-hour shifts and typically involve the monitoring of medium-sized areas. While many of the aforementioned techniques can be easily adapted, it is not feasible to simultaneously monitor an area carefully and repeatedly survey all the points of a graph whose nodes are sometimes hundreds of meters apart. \citet{chen2015designing}, however, focus on human patrols and a real city with the aim of establishing daily patrol routes that minimize the global average idleness of hotspots in a real environment. This work also does not yet address the joint restriction of finite time and effective surveillance time of an area, which involves being present at a specific node for a given period of time. 
    
    \item Other works focus on the surveillance of smaller environments, such as the floor of a building \citep[e.g.,][]{portugal2013applying, portugal2016cooperative, yan2016multi}, or are tested in artificial environments that could not be completely translated into a real case \citep[e.g.,][]{machado2002multicoor, almeida2003combining, santana2004multi, hwang2009cooperative, chen2023risk}. These approaches are therefore not directly comparable with the surveillance of a substantial part of a city. In addition, as is often the case, they do not include a differentiating factor between nodes or vertices in the environment. This differentiation is emphasized in our work by aiming to cover certain points while ignoring others with little interest in coverage.

\end{itemize}

\begin{landscape}
\setlength\tabcolsep{3pt} 
\begin{table}[h]
\caption{Main features of the literature review on patrol route design}
\bittiny 
    \begin{tabularx}{600pt}{llcclcccl}
        \hline
        \multirow{2}{*}{\textbf{Reference}} &  
        \multirow{2}{*}{\textbf{Objective}} &
        \multirow{2}{*}{\textbf{Agent}} &
        \multirow{2}{*}{\textbf{Environment} 
        } & 
        \multirow{2}{*}{\textbf{Model}} &
        \multirow{2}{*}{\textbf{Coordination}} &
        \multirow{2}{*}{\textbf{Area size} 
        } 
        & 
        \textbf{Preselected or} &
        \multirow{2}{*}{\textbf{Metrics}} 
        \\\vspace{-5mm}
        \\&&&&&&&\textbf{homogeneous nodes} 
        &\\
        \hline

        \multirow{2}{*}{\citet{machado2002multicoor}} &
        Minimize revisit time of all nodes &
        \multirow{2}{*}{—} & \multirow{2}{*}{Artificial} &  \multirow{2}{*}{Algorithms for}& \multirow{2}{*}{Both}& 
        \multirow{2}{*}{—} & \multirow{2}{*}{Yes} & \multirow{2}{*}{Idleness and exploration} 
        \\
        \vspace{-3.4mm}
        \\&&&& multi-agent systems&&&&time\\[1.3mm] 
        
        \citet{almeida2003combining} & Minimize revisit time of all nodes &—& Artificial & Control algorithm & Centralized &—& Yes & Idleness 
        \\[1.3mm] 

        \citet{santana2004multi} & Minimize revisit time of all nodes &—& Artificial & MARL & Distributed &—& Yes & Idleness 
        \\[1.3mm] 
        
        \multirow{2}{*}{\citet{hwang2009cooperative}} &
        Minimize total length path of all agents &
        \multirow{2}{*}{Robot} & \multirow{2}{*}{Artificial} &  \multirow{2}{*}{Cooperative auction}& \multirow{2}{*}{Distributed}& 
        \multirow{2}{*}{—} & \multirow{2}{*}{Yes} & \multirow{2}{*}{MINIMAX} 
        \\
        \vspace{-3.4mm}
        \\&and minimize revisit time of patrol points&&&system\\[1.3mm] 
        
        \citet{portugal2010msp} & Minimize revisit time of selected positions & Robot & Artificial & MSP algorithm & Centralized & Tiny & Yes & Average Node Frequency\\[1.3mm]   
        
        \citet{pasqualetti2012cooperative} &
        Minimize revisit time of selected positions&
        Robot & Real & Control algorithm & Centralized & Small & Yes &— 
        \\[1.3mm] 

        \citet{pasqualetti2012cooperative2} &
        Minimize revisit time of selected positions &
        Robot & Real &  Control algorithm & Centralized & 
        Small & Yes &—\\[1.3mm] 

        \multirow{2}{*}{\citet{STRANDERS201363}} &
        \multirow{2}{*}{Minimize revisit time of all nodes} &
        \multirow{2}{*}{Robot} & \multirow{2}{*}{Artificial} &  Divide and conquer & \multirow{2}{*}{Centralized} & 
        \multirow{2}{*}{—} & \multirow{2}{*}{Yes} & \multirow{2}{*}{Reward similar to idleness} 
        \\
        \vspace{-3.4mm}
        \\&&&&and greedy algorithms\\[1.3mm] 
        
        \citet{portugal2013applying} & Minimize revisit time of selected positions & Robot & Artificial & Bayesian learning & Distributed & Tiny & Yes & Idleness\\[1.3mm]   
        
        \citet{chen2015designing} & Minimize revisit time of hotspots & Human & Real & Ant colony & Distributed & Large & Yes & Idleness\\[1.3mm] 
        
        \citet{yan2016multi} & Minimize revisit time to selected positions& Robot & Real & Distributed algorithm & Distributed & Tiny & Yes & Idleness \\[1.3mm]
        
        \citet{portugal2016cooperative} & Minimize revisit time of selected positions & Robot & Real & Bayesian learning & Distributed & Tiny & Yes & Idleness\\[1.3mm]
        
        \citet{othmani2017multi} & Minimize revisit time of all nodes & Robot & Artificial & LSTM & Centralized &—& Yes & Idleness\\[1.3mm]  
        
        \citet{othmani2018} & Minimize revisit time of all nodes & Robot & Artificial & LSTM & Distributed &—& Yes & Idleness\\[1.3mm]
        
        \citet{othmani2019} & Minimize revisit time of all nodes & Robot & Artificial & LSTM & Distributed &—& Yes & Idleness\\[1.3mm]   
        
        \multirow{2}{*}{\citet{guo2023balancing}} &
        Balance between unpredictability in the routes&
        \multirow{2}{*}{Robot} & \multirow{2}{*}{Artificial} & \multirow{2}{*}{MARL} & \multirow{2}{*}{Distributed} & \multirow{2}{*}{—} & \multirow{2}{*}{Yes} & 
        \multirow{2}{*}{Idleness and time entropy}
        \\
        \vspace{-3.4mm}
        \\&and minimizing revisit time of selected positions\\[1.3mm] 

        \multirow{2}{*}{\citet{chen2023risk}} &
        Maximize coverage of hotspot&
        \multirow{2}{*}{Human} & \multirow{2}{*}{Artificial} & \multirow{2}{*}{Deep reinforcement} & \multirow{2}{*}{Distributed} & \multirow{2}{*}{Small} & \multirow{2}{*}{No} & 
        \multirow{2}{*}{Coverage}
        \\
        \vspace{-3.4mm}
        \\&roads and total roads&&&learning\\[1.3mm] 
        
        \hline
        \vspace{-1.5mm}
        \\
        \multirow{2}{*}{Our work} &
         Maximize crime surveillance during &
        \multirow{2}{*}{Human} & \multirow{2}{*}{Real} & 
        \multirow{2}{*}{MARL} & \multirow{2}{*}{Distributed} & \multirow{2}{*}{Medium} & \multirow{2}{*}{No} &  \multirow{2}{*}{Coverage index} 
        \\
        \vspace{-3.4mm}
        \\&a single shift \\[1mm] 
        \hline
    \end{tabularx}
\medskip
\parnotes
\label{table:literature_review}
\end{table}
\setlength\tabcolsep{6pt} 
\end{landscape}

Finally, to sum up, our proposal focuses on maximizing crime surveillance without the preselection of nodes to visit within a single work shift of human patrols. In our view, for optimal surveillance work, patrols should spend a relative amount of time at each location. In addition, they rarely can complete multiple cycles in an extensive area within the same working day. In this sense, \citet{chen2015designing} confirms that in order to achieve a significant decrease in the average node idleness, it is necessary to significantly increase the number of patrols in the system. Most other approaches are more suitable for other types of vigilance where the patrols have a smaller area to cover or when the number of patrols is enough to cover a determined area really well.

By assuming that surveillance time will not be infinite and that the generated route is not intended to be cyclical, discussing the idleness of the nodes becomes less relevant. This is because some nodes will not be visited due to time constraints or lack of interest, and certain conflict points may not warrant visitation on every occasion if they are too isolated from the other hotspots. For this reason, a new metric has been developed to measure the effectiveness of a generated route in terms of area coverage.

\section{Methodology}

\subsection{Definition of the Problem}

The patrolling problem in an urban environment can be classified as a \textit{decentralized partially observable Markov decision process} (dec-POMDP), also generalizable to \textit{multi-agent markov decision processes} (MMDPs), depending on whether the agents are allowed to know the entire state of the environment in their observations. In our model, the problem is formulated as a dec-POMDP with a tuple $<I, S, A, \Gamma, R, \Omega, O, \gamma>$ \citep{oliehoek2016concise}, where $I = \{a_1, a_2,\ldots, a_N\}$ is the society of $N$ patrolling agents; $S$ is the representation of the environment through which agents are able to move, represented as a set of states; $A$ is a function modeling the set of actions an agent can perform; $\Gamma$ is the set of conditional transitions between states, defined as $\Gamma(s_{t+1}|s_{t}, a_{t})$;  $R$ is the reward function, defined as $R(s_{t}, a_{t})$; $\Omega$ is the set of observations; $O$ is the probability of these observations; and $\gamma$ is the discount factor.

\subsection{The Environment Representation}

The monitoring area of our proposal could be an urban environment that is initially transformed into a grid and subsequently converted into an undirected graph. The urban space is thus parceled into nodes, connected among them in terms of their walkability. This approach is one of the most common options in the field of spatial representation of urban environments \citep{devia2013generating,birks2008synthesis}. Consequently, each node will represent a cell of the grid, with edges connecting neighboring cells that share a common road. The use of a grid as the basis of our graph also implies that the distance between the nodes in the real environment is the same, simplifying any consideration of choosing one path over another. 

Formally, the environment graph can be represented as a tuple $S=<V, E, C, \rho, \sigma>$, where $V=\{v_1,v_2,\ldots,v_M\}$ is the set of the $M$ nodes of the graph. $E:V \times V \rightarrow \{0,1\}$ is the \emph{mobility function}, which is also commutative, i.e., $E(v_i,v_j)=E(v_j,v_i)$. $E(v_i,v_j)=1$ if an agent can transit from the node $v_i$ to the node $v_j$ and vice versa; otherwise, $E(v_i,v_j)=0$. $C$ is the subset of nodes, $C \subseteq V$, that have to be monitored \footnote{It should be noted that there is no preselection of nodes within the designated area. However, auxiliary nodes that do not belong to the selected area are utilized to facilitate the movements of agents.}. 
$\rho_t:I \rightarrow V$ is the \emph{location function} relating each agent with their position at time $t$, and $\sigma:V \rightarrow \mathbb{R}$ a \emph{target function}. The higher the value of the target function for a node, the more important it is for agents to transit through it. Therefore, the goal of the model will be to maximize the coverage of those nodes with higher target values.  

With regard to the temporal constraints of the environment, the objective is to reproduce a shift of a group of human patrols (represented by the agents) at each episode $t$, $1 \leq t \leq T$, where $T$ is the total number of episodes of the simulation. Therefore, there is no penalty target at the conclusion of the episode $t$, i.e., agents are not rewarded for going faster. The number of steps that an agent can take in the environment is determined by the real surface that each node represents and by the time that we consider to be dedicated to monitoring each one of them. So, the maximum number of steps in the environment is something that is not fixed and depends on the area to be patrolled. We believe that for surveillance to be effective, it is necessary that a patrol remains in a given area for a certain period of time.

\subsection{Actions}

The actions in the model consist mainly of the movements of the agents. An agent can only move to any of the nodes directly connected to the one on which it is. In addition, it can remain on the same node. Let $\delta: V \times V \rightarrow \mathbb{N}^+$ be a \emph{distance function} between nodes that can be denoted as:

\begin{equation} \label{eq_distance_function}
\delta(v_i, v_j) = \begin{cases} 
                        0 &: v_i= v_j \\
                        1 &: E(v_i,v_j)=1\\
                        1+ \min\limits_{v_k \in V, v_k \neq v_i} \delta(v_k, v_j) &:E(v_i,v_j)=0 \land \exists v_k \in V, \\ 
                        & E(v_i, v_k) = 1  \land v_i \neq v_k
                    \end{cases}
\end{equation}

\noindent $\delta$ measures the distance between two nodes of the graph. If the nodes are adjacent, its value is $1$; otherwise, this function will calculate the shortest distance between these nodes. The action function of the model, $A: I \times V \rightarrow V $, can also be denoted, as can be seen in Equation \ref{eq_action_function}. 

\begin{equation} \label{eq_action_function}
A(a_i, v_j) = v_k, \delta(v_j, v_k)\le 1 
\end{equation}

\noindent The action function represents the movements that an agent can make. Thus, the agent can remain at its current node or move only to neighboring nodes, i.e., those with which it is connected via a path. 

\subsection{Observations}

Our model is designed for environments where agents may have partial information about the environment, i.e., we focus on problems with partially observable environments. Observations determine the information available to an agent when making a decision. The existence of a single type of agent in our model entails that all of them receive the same type of information to make a decision. The reason to withhold environmental information from patrols is largely in line with the argument put forward by \citet{santana2004multi}. Firstly, it simplifies the problem, accelerating the training process and facilitating the convergence of the model. Secondly, it sets up the environment more realistically as agents may not have complete information about its state. Third, it allows subsequent modifications of the training environment without compromising the size of the observation space. For these reasons, we introduce the concept of agents' \emph{line of sight} as a way to determine what information of the environment state is known by the patrols. As will be seen, this line of sight of the agents affects their training and thus the model performance in the resolution of the problem. A comparison of results has been sought by providing different numbers of patrols with different amounts of information.

\begin{equation} \label{eq_observations}
\begin{split}
    \Omega_t=<\{\rho_t(a_1),\rho_t(a_2), \ldots, \rho_t(a_N)\}, \\
    \{\upsilon_t(v_1), \upsilon_t(v_2), \ldots, \upsilon_t(v_M)\}, \\
    \{\sigma(v_1), \sigma(v_2), \ldots, \sigma(v_M)\} >   
\end{split} 
\end{equation}

The information contained in the observations, $\Omega_t$, at time $t$ (Equation \ref{eq_observations}), can be divided into three parts: (1) The locations, $\rho_t: I \rightarrow V$, of $N$ patrols in the model, represented by the node identifier where they are located at time $t$. (2) The number of visits agents have made to each one of the $M$ nodes in the line of sight, at time $t$, where $\upsilon_t: V \rightarrow \mathbb{N}^+$ is the function that computes this information. (3) The value of the target function, $\sigma$, of each node. The values calculated from $\upsilon_t$ are significant because the goal is to ensure that patrols monitor areas at a higher target value, with the intention that the value of revisiting an area in the short term will decrease once this area has been visited.


\subsection{Rewards} 

This problem falls within the scope of cooperative problem solving as the reward is not the same for all agents when performing a particular action. Moreover, the main difficulty of this problem lies in introducing a reward that allows the agent to adequately learn the task we want the model to perform. Unlike other reinforcement learning problems, where the agent clearly knows when it wins (it receives a reward) or loses (it does not receive a reward or receives a penalty), here there is no clear indicator that we can easily use to train the task of maximizing area coverage weighted by the target value of each area, which has been a challenge in the design of the reward function.

For this reason, we explored different approaches to the reward function in terms of its effect on agent training. First, we considered the sum of the objective values of all nodes monitored by the agents at each point in time, without taking into account the number of visits to a node. This resulted in all agents moving to a node with a high objective value and remaining there, since it was the node that gave them the highest reward. The goal of the problem, if we were to translate it to the real world, would be to coordinate the agents to search for the highest target value nodes in the environment rather than the optimal routes. Then, we modified that reward value by dividing it by the number of visits. This resulted in all agents moving to an area (a group of neighbor nodes) with a high target value and staying there. The two problems with this behavior are, firstly, that they may not mind passing many nodes outside the area to reach their target faster and, secondly, that they may not cover the most isolated hotspots in the model, resulting in certain areas being overwatched. Finally, the (cooperative) reward function we have used in our model is denoted in Equation \ref{eq_reward_function}. The function setting the final reward for taking an action at a particular moment will be the sum of the rewards of all agents at that moment, plus the added reward for the individual agent. This is aimed at minimizing the occurrence of lazy agents (those who wait for others to perform beneficial actions) among the group of agents.

\begin{equation} \label{eq_reward_function}
R_t(a_i) =R'_t(a_i) + \sum_{1\le j \le N} R'_t(a_j)
\end{equation}

\noindent An individual reward, defined as $R'_t(a_i)$ (equation \ref{eq_individual_reward_function}), introduces a penalty $\nu$, the \emph{coverage factor}, for those agents that pass through nodes either outside the surveillance area or of low interest, i.e., those whose rewards are less than 1. This penalty aims to keep the routes within the surveillance area or through areas of high interest as much as possible. Due to the introduction of penalties, a \emph{normalization factor} $\eta$ was added to the equation to partially harmonize rewards and penalties. Finally, an \emph{exploration function}, $\tau_t:V \rightarrow \mathbb{N}$ (Equation \ref{eq_exploration_function}), was defined to reward patrolling more nodes in the model and to allow a better exploration phase during training. In that equation, $\alpha^{+}$ is a reward parameter in our model, granted when the target value of a node exceeds a relevance threshold $\phi$, a parameter which determines the important nodes in the whole area. Conversely, if the relevance threshold is not met, $\alpha^{-}$ is given. These rewards are only granted when a node is visited for the first time without another patrol being present. This last addition is crucial for training since, as has been pointed out in other studies such as \cite{zhang2021centralized}, without it agents tend to stay in local minima of the problem and fail to identify targets that are difficult to reach.

\begin{equation} \label{eq_individual_reward_function}
R'_t(a_i) = 
\begin{cases}
\nu &: \rho_t (a_i) \notin C \\ \\


\frac{\sigma(\rho_t (a_i))}{\eta [ \upsilon(\rho_t (a_i))]} + \tau_t(\rho_t (a_i)) &: \rho_t (a_i) \in C \land \frac{\sigma(\rho_t (a_i))}{\eta [ \upsilon(\rho_t (a_i))]} \geq 1 \\ \\

\frac{\sigma(\rho_t (a_i)}{\eta [ \upsilon(\rho_t (a_i))]} + \tau_t(\rho_t (a_i)) + \frac{\nu}{2}&: \rho_t (a_i) \in C \land \frac{\sigma(\rho_t (a_i))}{\eta [ \upsilon(\rho_t (a_i))]} < 1 \\

\end{cases}
\end{equation}

\begin{equation} \label{eq_exploration_function}
\tau_t (a_i) = 
\begin{cases}
0 &: \upsilon(\rho_t (a_i)) \neq 1 \\

\alpha^{+} &: \upsilon(\rho_t (a_i)) = 1 \land \sigma(\rho_t (a_i)) \geq \phi \\ 

\alpha^{-} &: \upsilon(\rho_t (a_i)) = 1  \land \sigma(\rho_t (a_i)) < \phi \\ 

\end{cases}
\end{equation}

\subsection{Model Parameters}

The specific parameters used in the training of the model can be seen in Table~\ref{table:parameters}. Those not labeled as input parameters are calibrated by maximizing the reward function and minimizing the loss function. The first part of the table lists the general input parameters of the model, more specifically, the number of agents, the line of sight of each agent, and the strategy used to place each agent in the first node of the model that is being trained. 

The second section of the table shows all those parameters related to the reward function that have been explained in the previous section. Finally, note that the last column of the table contains the values used during the evaluation of the model, as will be explained in Section~\ref{sec_evaluation}.

\begin{table}[hbt!]
\caption{Model parameters}
\centering
\small
\begin{tabular}{llcc}
\multicolumn{4}{l}{\textit{\textbf{Input parameters}}} \\
\toprule
\textbf{Parameter} & \textbf{Meaning} & \textbf{Domain} & \textbf{Value} \\
\toprule
Zone ID & Identifier of the zone & $\mathbb{N}$ & \{3, 9, 10\} \\
\toprule
\# agents & Agents to be deployed & $\mathbb{N}$ & \{2, 5, 10\} \\
\toprule
Line of sight & Size of observation box & $\mathbb{N}$ & \{1, 3, 6\} \\
\toprule
Starting position & 
Initial distribution of the agents & Enum & \{Random, Best\} \\
\toprule
RL algorithm & Algorithm chosen to train the agents  & Enum &
\{VDPPO\} \\
\toprule
\\
\multicolumn{4}{l}{\textit{\textbf{Reward parameters}}} \\
\toprule
\textbf{Parameter} & \textbf{Meaning} & \textbf{Domain} & \textbf{Value} \\
\toprule
$\eta$ & Normalization factor & $\mathbb{R}$ &  10 \\
\toprule
$\phi$ & Relevance threshold & $\mathbb{R}$ & 10 \\
\toprule
$\nu$ & Coverage factor & $\mathbb{R}$ & \{-25, -10\} \\
\toprule
$\alpha^{-}$ & Exploration reward & $\mathbb{R}$ & \{5, 10\} \\
\toprule
$\alpha^{+}$ & Optimal exploration reward & $\mathbb{R}$ & \{50, 100\} \\
\toprule
\end{tabular}
\label{table:parameters}
\end{table}


\section{Urban Crime Surveillance Optimization}

The allocation of resources by public administrations is a matter of ongoing concern due to the lack of resources, and their misuse may directly impact the lives of citizens. In the case of the police, there is not always an adequate number of officers or resources available to carry out effective surveillance. Thus, it is necessary to prioritize surveillance in different areas of the city over others. In addition, it is also important to note in this domain that patrol routes are slightly different each day \citep{yin2012trusts, sherman2014integrated}. This is intended to prevent offenders from identifying surveillance patterns that they would learn to avoid in the real world.

\begin{figure}[h]
\centering
\includegraphics[width=14cm,keepaspectratio]
{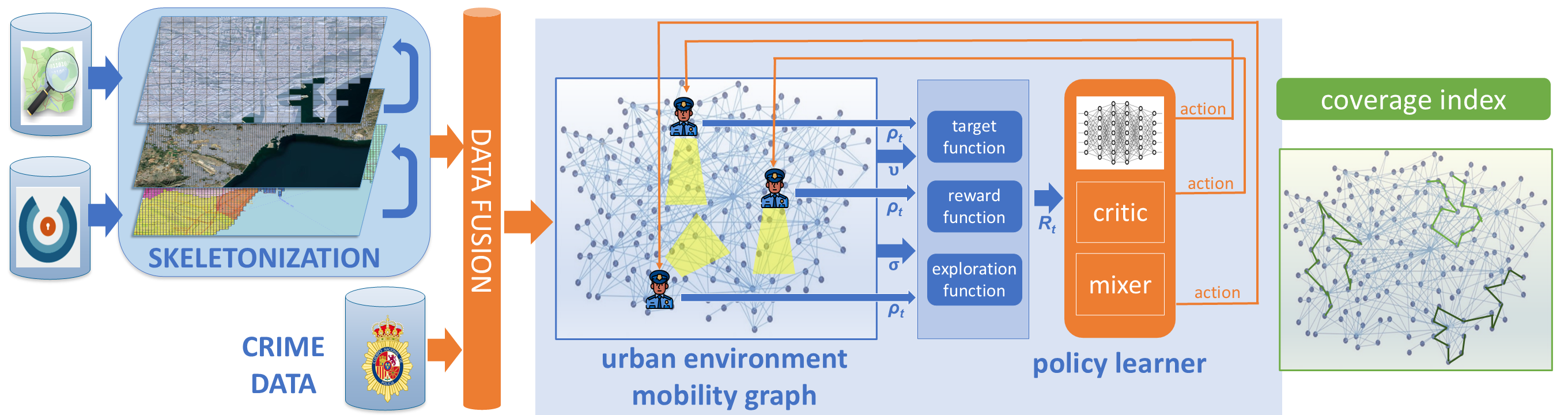}
\caption{Flowchart of the urban crime surveillance with our model.}
\label{fig:figure_diagram}
\end{figure}

Through this section, we particularize our model to the case of urban crime surveillance, and more concretely, to crime prevention in three areas of the city of Málaga (Spain) selected based on data availability. Two of these areas are similar in terms of crime rates and size, while the third area is larger and has a significantly lower crime density due to its size. The main goal is to collectively monitor the cells with the highest crime rate within that urban environment by setting different routes that complement each other for all police patrols. Figure~\ref{fig:figure_diagram} illustrates a flowchart describing all the stages of optimizing urban crime surveillance and thus how the methodology is applied to this problem. The input data to the flow are the geographical coordinates of the city's urban environment and information about its streets. This information is combined, and through a skeletonization process, the urban environment is digitized into a grid on which a graph showing the mobility between its cells is also overlaid. Next, the urban environment is then enriched with crime data that will shape the target function. In the second stage, the MARL algorithm (in our case, VDPPO, i.e., the one with the best performance) is used to learn the agents' policy. The results of the agent's movements in the environment are evaluated through the target and exploration functions, generating the reward that feeds the policy learner iteratively until the learning process converges. As a result of the whole process, the surveillance routes of the agents and the coverage index are generated. 

\subsection{The Urban Environment}

In order to create the environment, several input data have been used. Firstly, we collect information about the roads in the city of Málaga. This information is extracted and combined from two datasets: the road map of Andalusia, Spain, obtained from OpenStreetMap \citep{OpenStreetMap} and the list of the information on the roads of Málaga on the open data portal of Málaga City Council \citep{malagaSistemaInformacion}. 

Secondly, we collected various datasets that contain information on crime corresponding to citizen reports. These datasets were provided by the Spanish National Police Force and include all the crimes occurred in Málaga between 2010 and 2018, representing a total of 376,737 cases. These crime reports are manually introduced by the police officers and thus need to be geolocated in terms of the street name where they occurred. Eventually, only 304,125 crimes were successfully geolocated and used in this study.

\begin{figure}[h]
\centering
\includegraphics[width=9cm,keepaspectratio]
{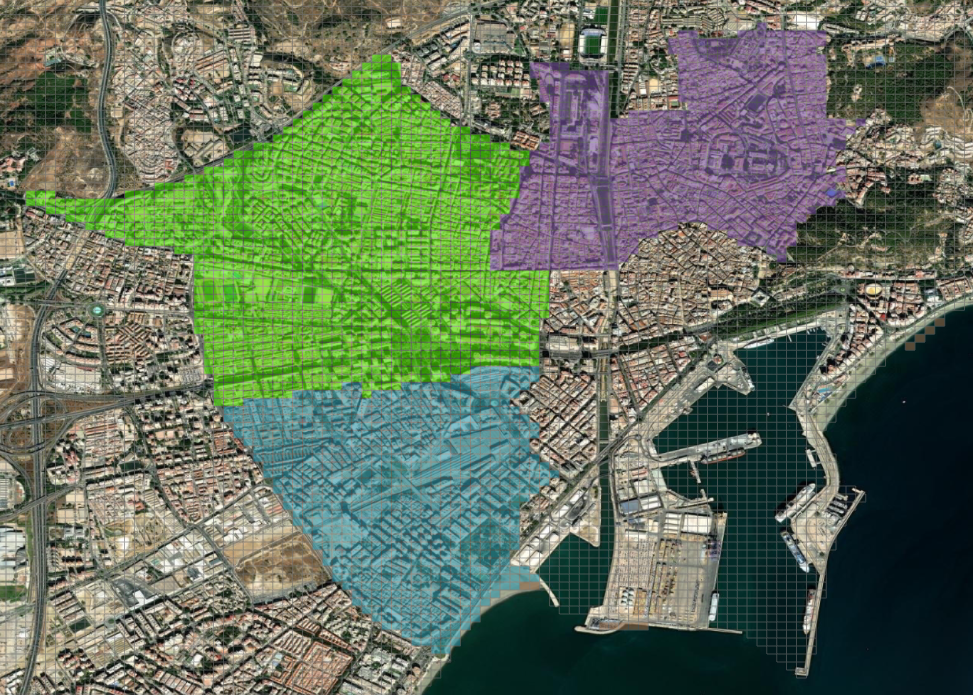}
\caption{Urban areas selected for patrol routing, Málaga, Spain: zones 3 (blue), 9 (purple), and 10 (green).}
\label{fig:figure_zones_model}
\end{figure}

After collecting and combining all the data, a grid was overlaid on the city map. All the information corresponding to streets and crimes was then added to each cell. In our case, the area of each cell corresponds to 2,500 m\textsuperscript{2}, 50 meters per side, and connections between cells correspond to the roads between them. These cell dimensions were chosen in this case to provide a realistic approximation of the movement of an agent, preventing it from taking too long to move from one cell to another. Additionally, this size contributes to a better and more realistic distribution of criminal activity, which with larger grid sizes may be condensed in a single node. Note also that, to our knowledge, our grid size is one of the smallest that can be found in the literature of crime prediction or hotspots analysis. For example, \citet{KADAR2019107}  and \citet{RUMMENS2017255} use a cell size of 200 meters per side, \citet{adepeju2016novel} 250 meters per side, and \citet{lee2020theory} 152 meters per side.

\begin{figure}[h]
\centering
\includegraphics[width=13.5cm,keepaspectratio]{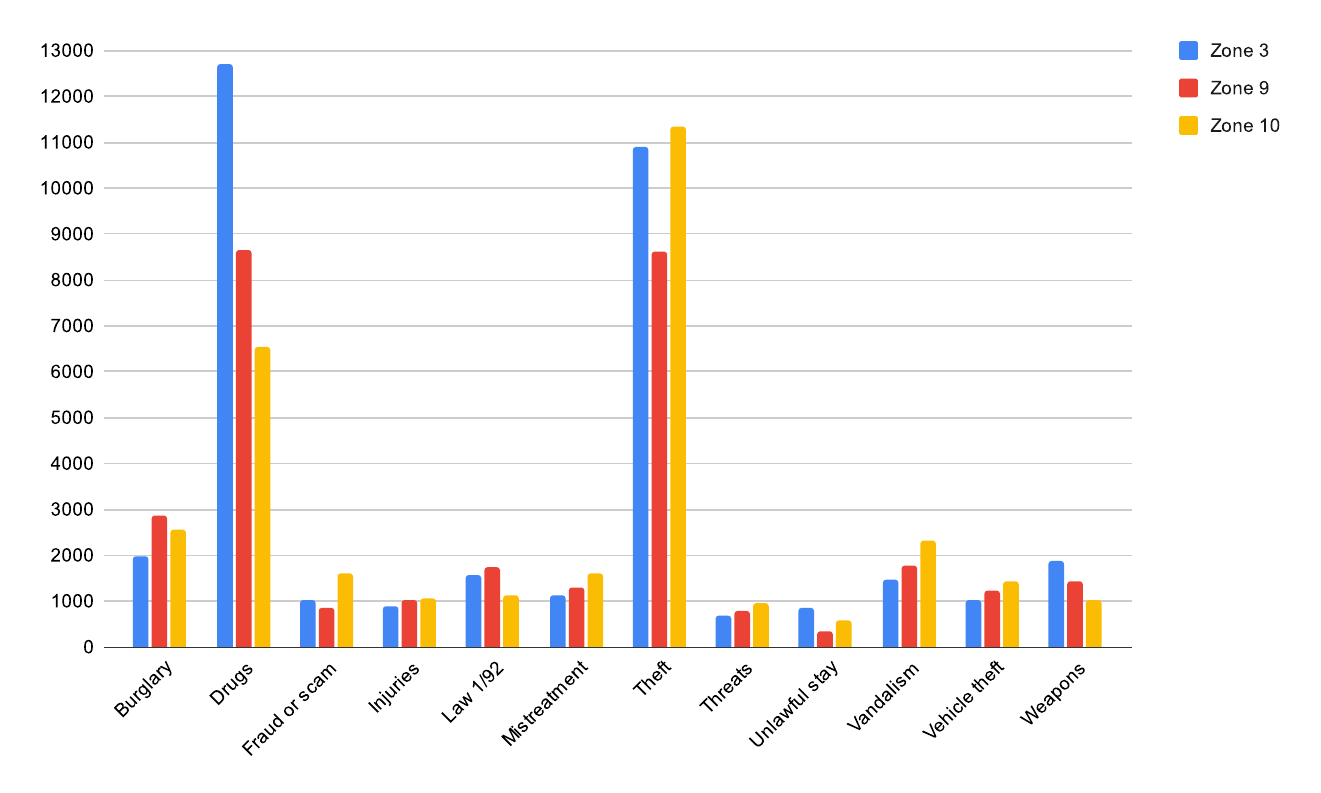}
\caption{Number of crimes by typology in each studied area.}
\label{fig:figure_crime_areas}
\end{figure}

From the grid, the area of the city to be patrolled had to be selected. In this case, we focused on three urban areas in the city center (Figure~\ref{fig:figure_zones_model}). The first two areas have an approximate size of 1.8 km\textsuperscript{2} and a crime density of 1,999 and 1,672 offenses per year per km\textsuperscript{2} respectively, while the third area is 3 km\textsuperscript{2} with a crime density of 1,080 offenses per year and per km\textsuperscript{2}. The areas were chosen because they are all residential, although there are differences in the density of crime, as well as in its typology, as can be seen in Figure~\ref{fig:figure_crime_areas}. The histogram shows that the most common criminal activities in the city are drug-related crimes and thefts, which are most frequently observed in all three zones. Zone 3 encompasses the city's main railway station, which has become a focus for drug dealing and theft, largely due to the high volume of passengers in the area. This area has the highest crime rate of the three areas surveyed. In contrast, zone 10 is considerably larger than the other two and is mainly residential, with no discernible presence of tourists. The size of the zone translates into crime levels that make it comparable to the other two, and the zone is notable for the total number of reported burglaries. Finally, zone 9, comparable in size to zone 3, is also residential, like zone 10, but is located closer to the city center, resulting in a lower but comparable crime rate to zone 3. Nevertheless, it is not the main hotspot for any type of criminal activity.

Additionally, to ensure that the selection of a particular area has not disrupted the usual flow of people by excluding any connections between the cells in the area that is crossed by external roads, additional cells outside the zone were added to reproduce a perfect rectangle. Finally, every cell not containing at least one road was removed due to its inaccessibility, and each remaining cell was also provided with its own crime rate, which will be our target function $\sigma$.

\begin{figure}[h]
\centering
\includegraphics[width=4cm,keepaspectratio]{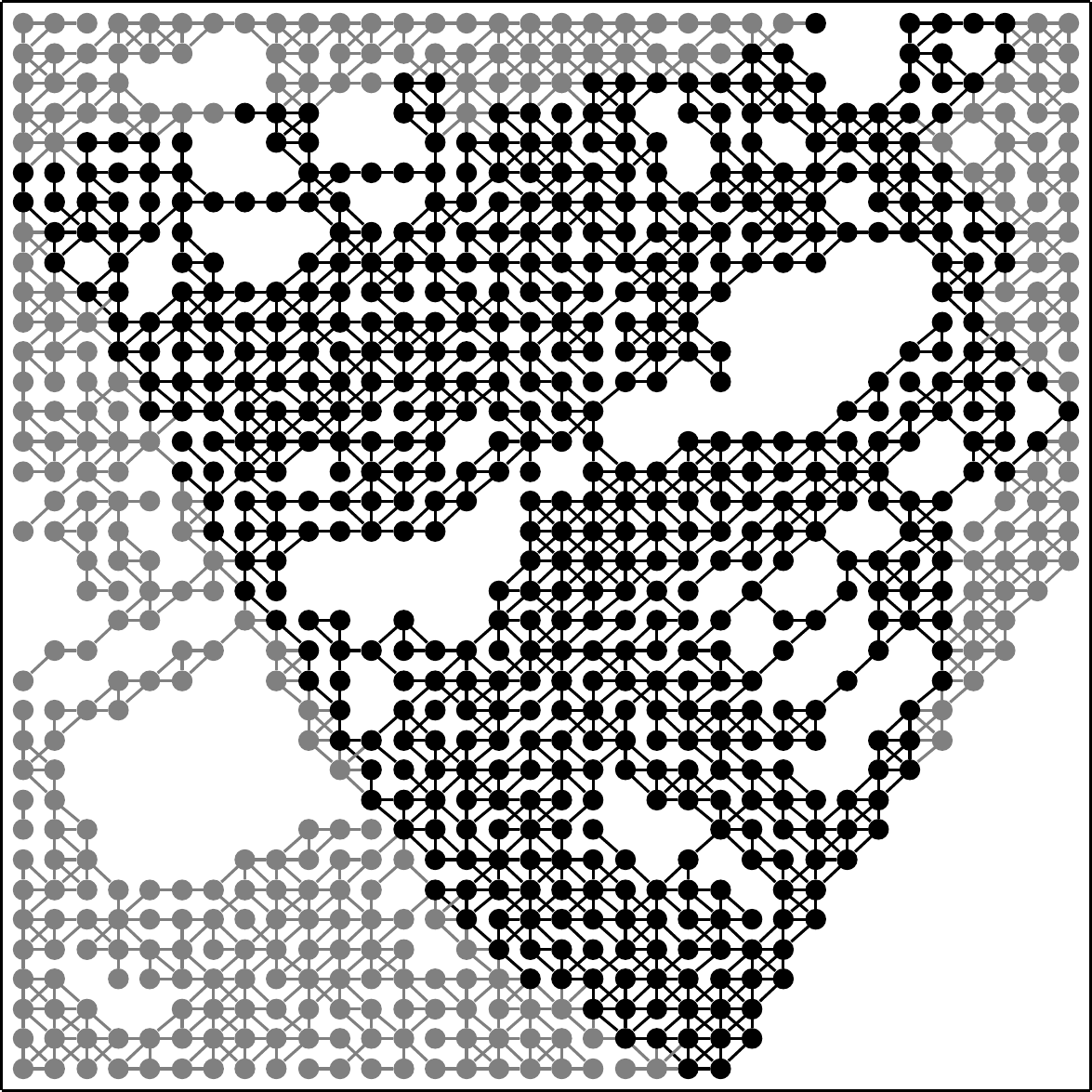}
\includegraphics[width=4cm,keepaspectratio]{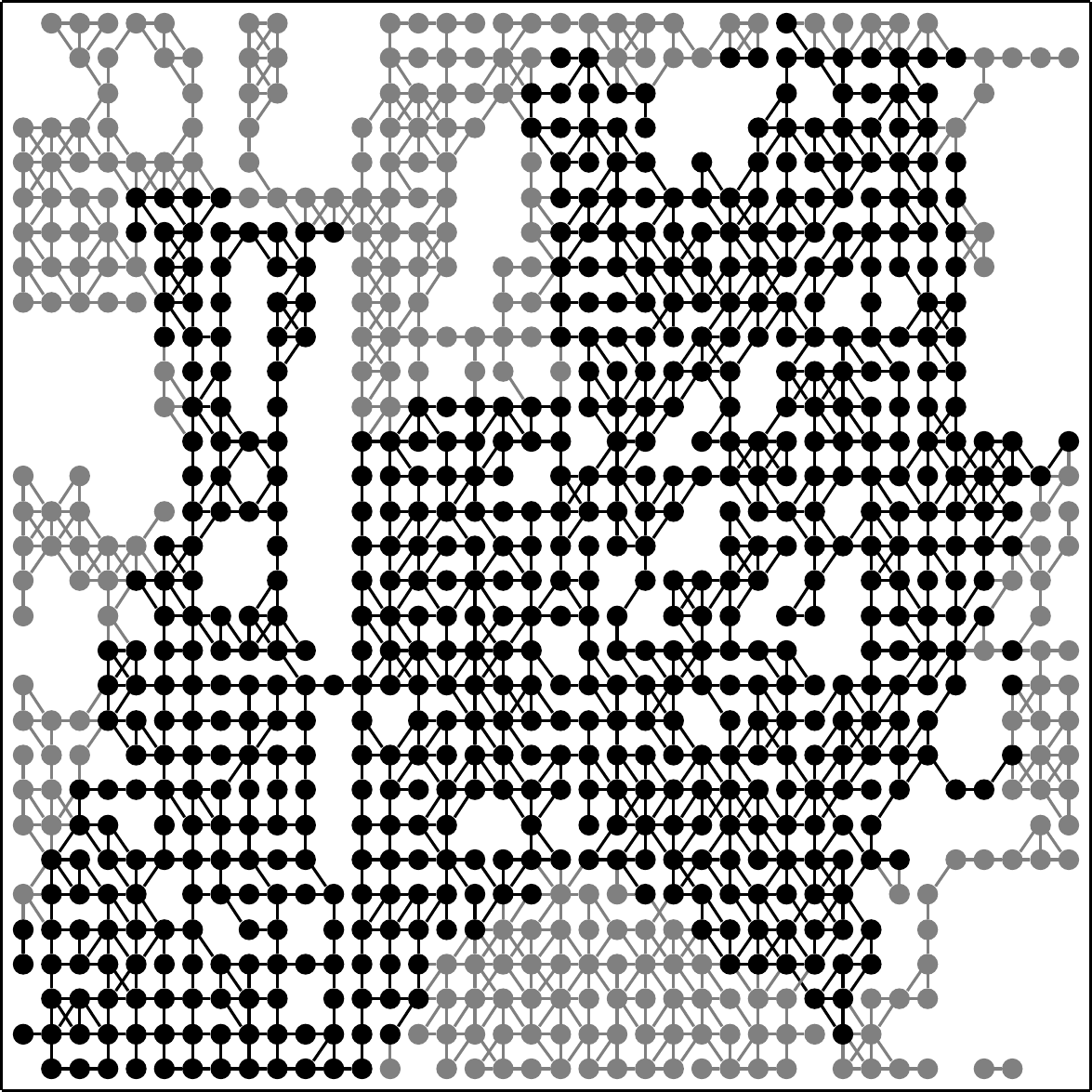}
\includegraphics[width=4cm,keepaspectratio]{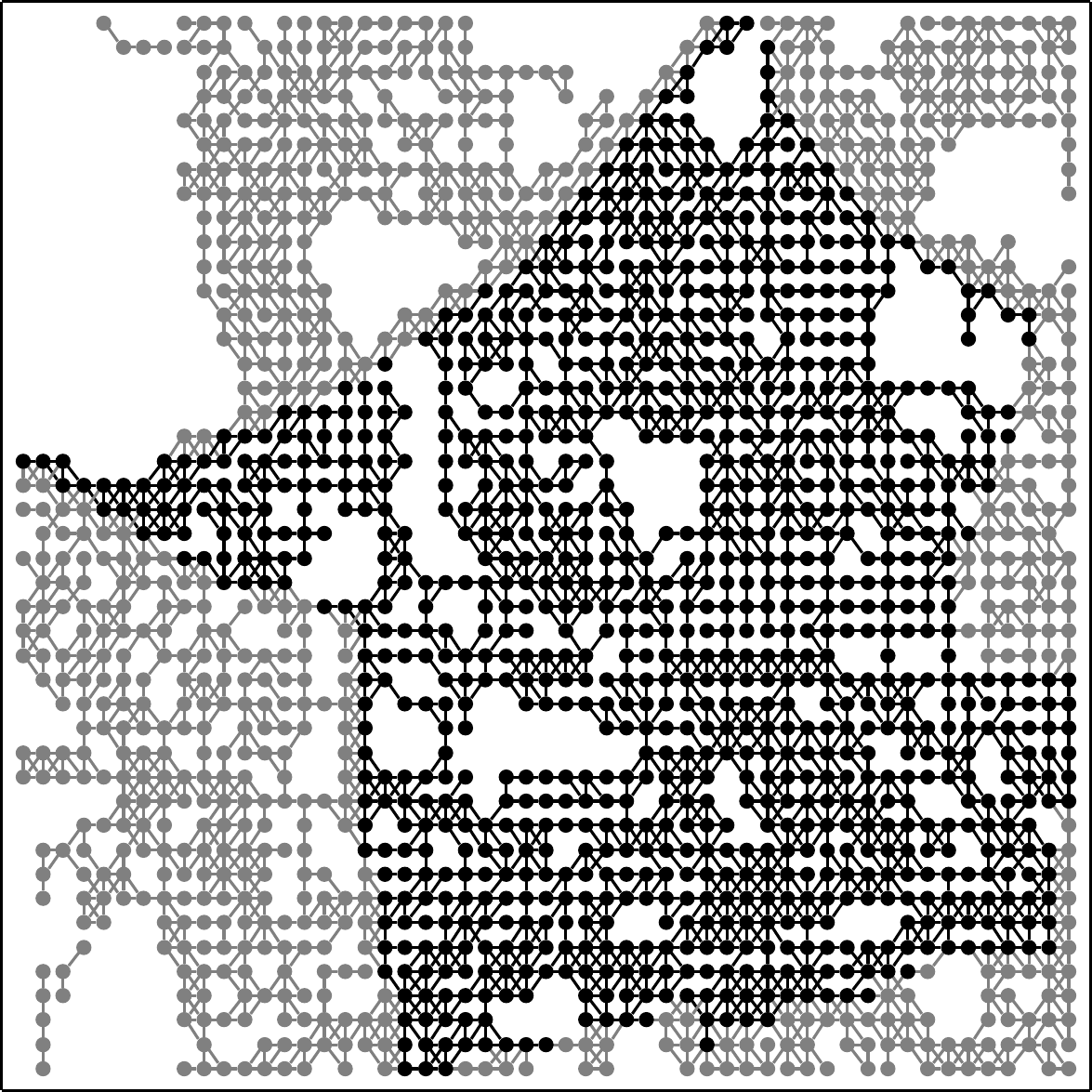}
\caption{Graphs generated from the selected areas in Figure \ref{fig:figure_zones_model}: zone 3 (left), 9 (center), and 10 (right).}
\label{fig:figure_graphs}
\end{figure}

After all this processing, each zone grid was converted into a graph, as illustrated in Figure~\ref{fig:figure_graphs}. These graphs, although derived from a grid, become irregular due to the removal of non-navigable cells, resulting in large gaps or voids in the graphs. For instance, in zone 3, there are two main gaps: the Mediterranean Sea in the bottom right corner and the railway tracks in the middle; and in zone 10, there are hills in the top left corner. Table~\ref{table:characteristics_areas} shows the main characteristics of each patrolling area, i.e., its dimensions, annual crime rates, and the number of nodes and edges of the graph generated from the grid.

\begin{table}[h]
\caption{Characteristics of each patrolling area.}
\centering
\small 
    \begin{tabularx}{287pt}{ccccc}
        \toprule
        \textbf{Zone Id.} & \textbf{Extension} & \textbf{Annual crime rate} & \textbf{\# nodes} & \textbf{\# edges}\\
        \midrule
        zone 3 & 1.81 km\textsuperscript{2} & 1,999 crimes / km\textsuperscript{2} & 921 & 2,095 \\ 
        zone 9 & 1.83 km\textsuperscript{2} & 1,672 crimes / km\textsuperscript{2} & 923 & 1,809 \\ 
        zone 10 & 2.98 km\textsuperscript{2} & 1,080 crimes / km\textsuperscript{2} & 1766 & 3,733 \\ 
        \hline
    \end{tabularx}
\medskip
\label{table:characteristics_areas}
\end{table}

Once the environments were designed and configured, three main parameters needed to be configured: the number of agents, their initial positions, and their range of vision (i.e., \textit{line of sight}). Regarding the number of agents, we explored a small range of values, all of them feasible according to the law enforcement resources. This range allows for adequate coverage of the environment while also being achievable with the city’s police resources. Similarly, a range of values was used for the agents' \textit{line of sight}. Finally, for the initial agent positions, two possibilities were tested: randomly positioning the patrols and deploying them at the graph nodes with the highest target function ($\sigma$) value.

\subsection{Assumptions}

When representing real-world scenarios using reinforcement learning models, it is essential to make clear the constraints and assumptions about the real problem to be considered. The more complex the problem to be reproduced, the more difficult it will be to define an effective reward function that facilitates an adequate model learning process. Conversely, if the problem is oversimplified, there is a risk that the results of model training may not be applicable (or may be meaningless) in real situations. For this reason, the following assumptions have been made to model this police patrolling scenario:

\begin{itemize}
    \item Agents are not constrained to pass through any particular cell during the surveillance, meaning there is no requirement to monitor any specific cell within the area on the map.
    \item Continuous over-surveillance of an area reduces the effectiveness of patrols in that same area for a short period \citep{eck1993threat}.
    \item Patrols will not have to dynamically deviate from their route. To establish a route, we assume that no supervening factors will force agents to change course. Although this does not fully conform to real-world conditions, we believe that, in most cases, patrols can continue from where they left off without significantly affecting the resolution of the problem.
    \item All patrols will be equally effective in crime surveillance and will behave similarly, meaning the agents will be homogeneous or interchangeable and, therefore, will share a common policy.
    \item Each episode of time $t$ will be approximately equivalent to 10 minutes. As a result, a shift of 8 hours would be equivalent to 48 steps, which we have rounded to 50 steps.
\end{itemize}

\begin{figure}[h]
\centering
\includegraphics[width=8cm,height=5cm,keepaspectratio]{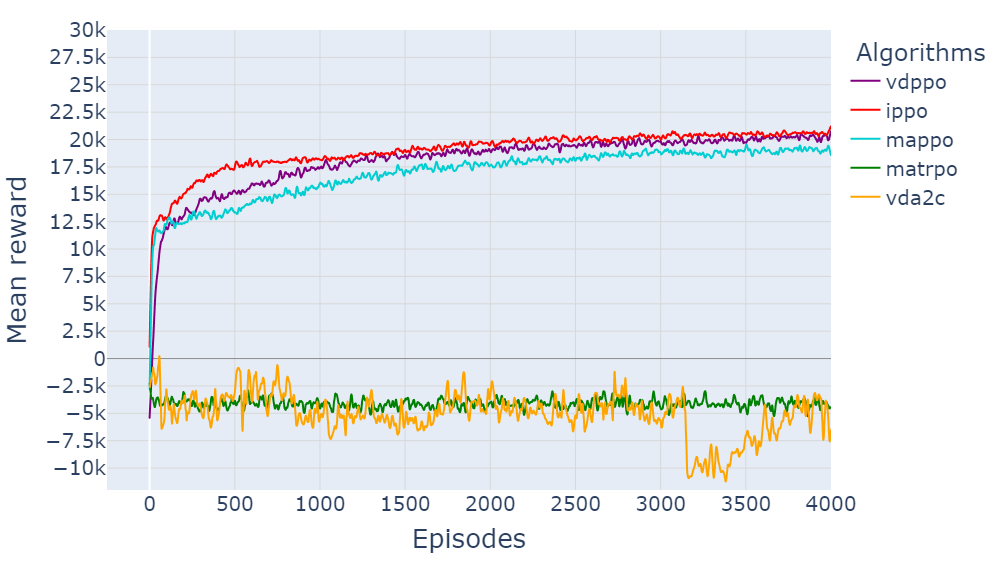}
\caption{Comparison of the maximum rewards obtained with IPPO (red), MAPPO (cyan), VDPPO (purple), VD2AC (yellow), MATPRO (green), using a line of sight of 3.}
\label{fig:failed_train_matrpo_vd2ac}
\end{figure}

To address this police patrol optimization problem, various MARL algorithms were explored. Algorithms that extend or implement \textit{Deep Deterministic Policy Gradient} (DDPG), such as Independent DDPG (IDDPG) \citep{lillicrap2015continuous}, Multi-Agent DDPG (MADDPG) \citep{lowe2017multi}, or \textit{factored multi-agent centralised policy gradients} (FACMAC) \citep{peng2021facmac}, are designed for continuous action space environments and are not applicable to discrete action space environments, such as our scenario. We also tested the algorithms belonging to the \textit{Advantage Actor Critic} (A2C) family, such as \textit{Multi-Agent A2C} (MAA2C) \citep{iqbal2019actor} or \textit{Value Decomposition A2C} (VDA2C) \citep{su2021value}, and the \textit{Trust Region Policy Optimizer} (TRPO) family, such as \textit{Multi-Agent TRPO} (MATRPO) \citep{li2023multiagent} or \textit{Heterogeneous-Agent Multi-Agent TRPO} (HATRPO) \citep{kuba2021trust}. However, they were discarded because they were unable to learn to solve the problem (see Figure~\ref{fig:failed_train_matrpo_vd2ac}). The three algorithms exhibiting the best performance were: \textit{Independent PPO} (IPPO) \citep{de2020independent}, an adaptation of PPO defined for independent learning; \textit{Multi-Agent PPO} (MAPPO) \citep{yu2022surprising}, a multi-agent adaptation of PPO derived from IPPO; and \textit{Value Decomposition PPO} (VDPPO) \citep{ma2022value}, an extension of IPPO focusing on credit assignment learning. 

In all cases, the policy has been shared among the group of agents, as they are homogeneous agents. The architecture of the learning network used was GRU (\emph{Gated Recurrent Unit}) \citep{cho2014learning} with an encoding layer of 256–256 neurons. This architecture is similar to LSTM  \citep{hochreiter1997long} but provides a simpler structure \citep{abbaspour2020comparative}. Note also that the implementation of the algorithms used in our work is that provided by the MARLlib library \citep{hu2022marllib}, which, in turn, encapsulates the functionality of the RLlib library from Ray \citep{liang2018rllib}.

\section{Evaluation}\label{sec_evaluation}

As can be seen in the last column of Table~\ref{table:parameters}, the number of patrols used in the evaluation was set at 2, 5, and 10, all these values within the resources available to the police. The \textit{line of sight} range values of 1, 3, and 6 were selected to represent varying levels of information available to the model. Finally, zone 10 differs from the others in terms of crime dispersion and size, requiring specific adjustments to train the model effectively. To address these training challenges, the coverage factor for this zone was increased from -25 to -10, and the exploration rewards were doubled to encourage agents to explore the environment thoroughly. These adjustments were necessary because, without increasing the exploration value and reducing penalties, the model had difficulty to train.

The reinforcement learning algorithm used in the evaluation was VDPPO. It is noteworthy that both IPPO and MAPPO were also able to complete the training process and achieve the desired outcomes. However, IPPO required three times longer to train than VDPPO, with an average training time of between eight and twelve hours per model. Additionally, MAPPO consistently performed slightly worse than the IPPO and VDPPO algorithms. Regarding the specific parameters of the VDPPO algorithm (Table~\ref{table:hyperparameters}), a hyperparameter search was conducted to identify the most suitable values within typical ranges. Further information on the parameters, including their meanings and explanations, can be found in \citet{schulman2015high, schulman2017proximal, albrecht2024multi}.

\begin{table}[hbt!]
\caption{VDPPO hyperparameters}
\centering
\small
\begin{tabular}{llcc}
\textbf{Parameter} & \textbf{Meaning} & \textbf{Domain} & \textbf{Value} \\
\toprule
lr & Learning rate & $\mathbb{R}$ & 0.0005 \\
\toprule
$\lambda$ & Lambda & $\mathbb{R}$ & 0.95 \\
\toprule
entropy coeff & Entropy coefficient & $\mathbb{R}$ & 0.01 \\
\toprule
GAE & Use of Generalized Advantage Estimator & \textit{Boolean} & True \\
\toprule
kl coeff & Initial coefficient for KL divergence & $\mathbb{R}$ & 0.3 \\
\toprule
\end{tabular}
\label{table:hyperparameters}
\end{table}

To evaluate the agents' learning performance, the value of the reward function was initially studied, as well as the effect of the line of sight parameter and the number of patrols deployed on the simulations. As expected, the greater the amount of information about the state of the environment that agents receive, the better the results. Also, we wanted to study the difference in performance depending on the initial node of patrol deployment, i.e., whether patrols are initially deployed in high-crime nodes or, on the contrary, in random positions.

Besides the information provided by the rewards and the loss function of the MARL problem, it is important to check the behavior of the agents to ensure that they have learned the expected behavior and that there is no flaw in the problem setup that causes them to behave in an unexpected way in order to maximize the reward they receive. The only two ways to achieve this are either to have a problem that is simple enough in terms of objectives to be able to define a reward function that simultaneously converges on and perfectly represents the realistic score, or, as in the case of this paper, to set up several metrics to check that, once the model is trained, it behaves as expected.

To study the results, we have not found clear metrics in the literature to help determine which solutions are effective for crime surveillance, mainly because defining what constitutes effective surveillance is not straightforward. In multi-agent patrolling problems, the most used performance criterion is idleness: the time elapsed between consecutive visits to pre-established observation points \citep{huang2019survey}. Our model does not preset routes but rather learns or designs coordinated routes that cover the areas with the highest crime rates. So, from our perspective, the idleness criterion was not appropriate to our approach. For this reason, two metrics have been established in order to study the results. The first and simplest is the entropy, which will help us assess how random the routes have been over the 100 runs and especially whether random deployment compensates in terms of randomness compared to the initial deployment in the best nodes. The second metric has been defined to evaluate the coverage of the environment. For this purpose, we define the \emph{coverage index}, an indicator inspired by the \textit{predictive accuracy index} (PAI) \citep{chainey2008utility}, a well-known metric within the field of criminology, specifically focusing on crime hotspot detection. The PAI is used to measure the good performance of predictive models of crime. PAI evaluates the quality of prediction of hotspots on a map in terms of a minimum threshold of surface coverage. The accuracy is measured as the proportion of predicted crimes with respect to the chosen coverage area. The values of this area are typically in the range from $3\%$ to $20\%$. 
\\
\\
\textbf{Coverage index ($|W_{\psi}|$)}: This is a measurement indicator that we have designed based on the coordinated coverage of a surveillance area. Since we have information about crimes that have occurred in these areas, we rank all nodes in the area to be covered by the number of crimes and extract the number of nodes that represent the chosen percentage, $\psi$, within the range of $3\%$ to $20\%$. What is measured is the number of nodes with the highest criminal rate, covered by the coordinated routes of the agents. This allows us to evaluate the effectiveness of the selected routes in terms of the coverage of the $\psi$ percentage of nodes with the highest crime rates. Let $G$ be the subset of nodes, $G \subseteq C$, to be monitored, covered in the routes of the agents, and $Z \subseteq G$, a subset that fulfills the following conditions (Equations \ref{eq_coverage_index_1} and \ref{eq_coverage_index_2}):

\begin{equation} \label{eq_coverage_index_1}
\forall v_x \in Z, \forall v_y \notin Z \land v_y \in G, \sigma(v_x) > \sigma(v_y) \\
\end{equation}

\begin{equation} \label{eq_coverage_index_2}
|Z| = \frac{\psi |G|}{100} 
\end{equation}

\noindent Equation \ref{eq_coverage_index_1} expresses that all nodes in the subset $Z$ have the highest values of the target function, $\sigma$, with respect to $G$, which is the set of all nodes included in the agents' routes. Equation \ref{eq_coverage_index_2} states that the cardinality of $Z$ is determined by $\psi$, having $Z$ therefore the percentage $\psi$ of nodes with the highest values of $\sigma$. Finally, the coverage index, which depends on the value of $\psi$, is the cardinality of the subset $W_{\psi}$ with the nodes in $Z$ visited by any agent in any simulation episode (Equation \ref{eq_coverage_index_3}). 

\begin{equation} \label{eq_coverage_index_3}
W_{\psi}=\{v_i \in Z | \exists t, 1 \leq t \leq T,\exists a_j \in I, \rho_t (a_j) = v_i \}
\end{equation}


\subsection{Comparative Study}

As mentioned above, our model seeks an optimal selection of coordinated paths by maximizing a target function within a limited time and without having a fixed destination. This means that we cannot compare our results with the state-of-the-art baseline algorithm, CBLS \citep{portugal2013applying, portugal2016cooperative, guo2023balancing}, because it works based on the concept of idleness and assumes that all nodes have to be visited. In our problem, it would be impossible to visit all nodes with our limited resources. For this reason, we have developed a greedy algorithm that we believe is comparable to our solution except in terms of communication. This greedy algorithm allows us to determine the minimum threshold of the target function that can be achieved. In this manner, each agent will move to the neighboring cell with the highest score within its reach. All agents will make their decisions simultaneously, similar to the MARL models. 

For this greedy algorithm, there is no point in discussing entropy or unpredictability in the case of a fixed initial position of the patrols, such as deployment at the optimal location, because this model is deterministic and, therefore, will always produce the same route with the same starting point. In the case of an initial random position, it will generate something similar; however, if these positions are very close to each other, they may influence the generated routes by moving toward the same cells simultaneously or by one patrol reaching a position before another that was also heading there. This is further exacerbated if the initial position places the patrols in a location that is not particularly relevant on the map.

\begin{figure}[hbt!]
\centering
\includegraphics[width=14cm,height=6cm,keepaspectratio]{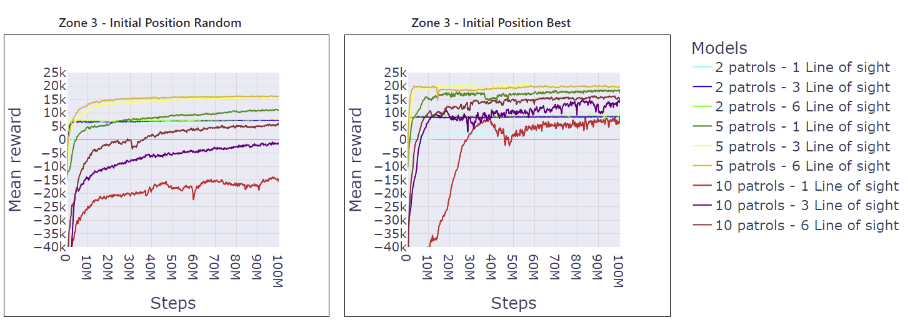}
\caption{Comparison of model training in zone 3 in terms of mean reward}
\label{fig:UnionGraficasArea3}
\includegraphics[width=14cm,height=6cm,keepaspectratio]{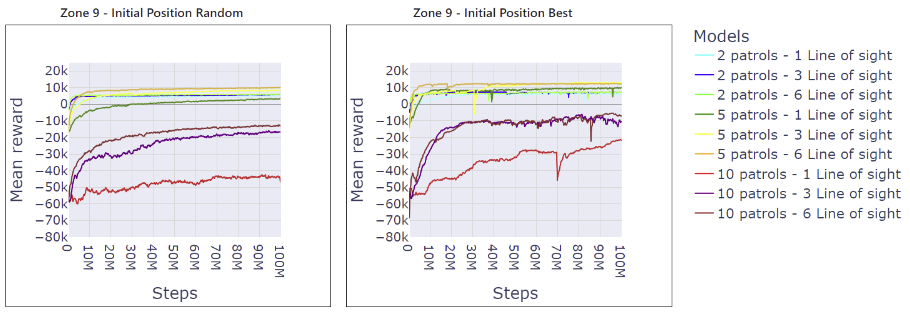}
\caption{Comparison of model training in zone 9 in terms of mean reward}
\label{fig:UnionGraficasArea9}
\includegraphics[width=14cm,height=6cm,keepaspectratio]{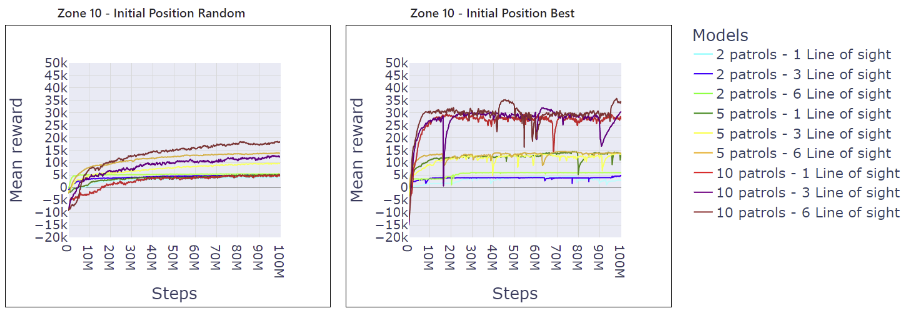}
\caption{Comparison of model training in zone 10 in terms of mean reward}
\label{fig:UnionGraficasArea10}
\end{figure}

\begin{table}[hbt!]
\caption{Results in zone 3 in terms of coverage index (for values of $\psi$ 3\%, 5\%, 10\%, and 20\%) and entropy}
\centering
\footnotesize
    \begin{tabularx}{330pt}{cccccccc}
        \toprule
        \textbf{Line of sight} & \textbf{Initial position} & \textbf{\# patrols} & \textbf{$|W_{3}|$} & \textbf{$|W_{5}|$} & \textbf{$|W_{10}|$} & \textbf{$|W_{20}|$} & Entropy \\
        \midrule
        Greedy & random & 2 & 0.284 & 0.293 & 0.279 & 0.239 & 5.21 \\
        Greedy & random & 5 & 0.512 & 0.557 & 0.550 & 0.483 & 5.45 \\
        Greedy & random & 10 & 0.718 & 0.762 & 0.767 & 0.697 & 5.62 \\
        Greedy & best & 2 & 0.760 & 0.580 & 0.350 & 0.180 & 0.00 \\
        Greedy & best & 5 & 0.760 & 0.620 & 0.490 & 0.310 & 0.00 \\
        Greedy & best & 10 & 0.820 & 0.690 & 0.570 & 0.440 & 0.00 \\
        \midrule
        1 & random & 2 & 0.942 & 0.758 & 0.633 & 0.447 & 4.83 \\
        1 & random & 5 & 0.967 & 0.895 & 0.773 & 0.592 & 5.26 \\
        1 & random & 10 & 0.938 & 0.819 & 0.732 & 0.574 & 5.21 \\
        1 & best & 2 & 0.937 & 0.893 & 0.748 & 0.492 & 4.2 \\
        1 & best & 5 & 0.999 & 0.971 & 0.889 & 0.735 & 5.16 \\
        1 & best & 10 & \textbf{1.000} & 0.992 & 0.928 & 0.850 & \textbf{5.79} \\
        \midrule
        3 & random & 2 & 0.929 & 0.756 & 0.627 & 0.445 & 4.77 \\
        3 & random & 5 & 0.978 & 0.95 & 0.831 & 0.674 & 5.22 \\
        3 & random & 10 & 0.975 & 0.924 & 0.823 & 0.708 & 5.59 \\
        3 & best & 2 & 0.928 & 0.907 & 0.769 & 0.501 & 4.23 \\
        3 & best & 5 & 0.999 & 0.984 & 0.909 & 0.783 & 5.06 \\
        3 & best & 10 & \textbf{1.000} & 0.995 & \textbf{0.966} & 0.890 & 5.78 \\
        \midrule
        6 & random & 2 & 0.908 & 0.747 & 0.625 & 0.459 & 4.91 \\
        6 & random & 5 & 0.977 & 0.938 & 0.83 & 0.675 & 5.21 \\
        6 & random & 10 & 0.996 & 0.968 & 0.894 & 0.772 & 5.52 \\
        6 & best & 2 & 0.868 & 0.871 & 0.769 & 0.551 & 4.48 \\
        6 & best & 5 & 0.999 & 0.982 & 0.903 & 0.764 & 4.97 \\
        6 & best & 10 & 0.998 & \textbf{0.999} & 0.994 & \textbf{0.916} & 5.71 \\
        \hline
    \end{tabularx}
\medskip
\label{table:results_zone_3}
\end{table}

\begin{table}[hbt!]
\caption{Results in zone 9 in terms of coverage index (for values of $\psi$ 3\%, 5\%, 10\%, and 20\%) and entropy}
\centering
\footnotesize
    \begin{tabularx}{330pt}{cccccccc}
        \toprule
        \textbf{Line of sight} & \textbf{Initial position} & \textbf{\# patrols} &\textbf{$|W_{3}|$} & \textbf{$|W_{5}|$} & \textbf{$|W_{10}|$} & \textbf{$|W_{20}|$} & Entropy \\
        \midrule
        Greedy & random & 2 & 0.232 & 0.197 & 0.189 & 0.179 & 5.42 \\
        Greedy & random & 5 & 0.496 & 0.410 & 0.381 & 0.345 & 5.60 \\
        Greedy & random & 10 & 0.664 & 0.591 & 0.578 & 0.535 & 5.76 \\
        Greedy & best & 2 & 0.570 & 0.430 & 0.290 & 0.170 & 0.00 \\
        Greedy & best & 5 & 0.630 & 0.460 & 0.310 & 0.200 & 0.00 \\
        Greedy & best & 10 & 0.890 & 0.680 & 0.510 & 0.350 & 0.00 \\
        \midrule
        1 & random & 2 & 0.784 & 0.728 & 0.652 & 0.416 & 5.11 \\
        1 & random & 5 & 0.851 & 0.797 & 0.689 & 0.519 & 5.54 \\
        1 & random & 10 & 0.389 & 0.339 & 0.373 & 0.372 & 5.6 \\
        1 & best & 2 & 0.811 & 0.721 & 0.658 & 0.437 & 4.41 \\
        1 & best & 5 & 0.941 & 0.872 & 0.888 & 0.692 & 5.53 \\
        1 & best & 10 & 0.900 & 0.808 & 0.825 & 0.654 & 5.44 \\
        \midrule
        3 & random & 2 & 0.801 & 0.735 & 0.714 & 0.441 & 4.96 \\
        3 & random & 5 & 0.918 & 0.894 & 0.863 & 0.622 & 5.37 \\
        3 & random & 10 & 0.925 & 0.884 & 0.837 & 0.678 & 5.72 \\
        3 & best & 2 & 0.894 & 0.838 & 0.823 & 0.524 & 4.49 \\
        3 & best & 5 & \textbf{1.000} & \textbf{0.999} & \textbf{0.997} & 0.782 & 5.32 \\
        3 & best & 10 & 0.987 & 0.961 & 0.939 & 0.764 & 5.92 \\
        \midrule
        6 & random & 2 & 0.816 & 0.747 & 0.711 & 0.439 & 5.02 \\
        6 & random & 5 & 0.965 & 0.948 & 0.916 & 0.662 & 5.36 \\
        6 & random & 10 & 0.958 & 0.926 & 0.895 & 0.731 & 5.75 \\
        6 & best & 2 & 0.933 & 0.830 & 0.785 & 0.480 & 4.42 \\ 
        6 & best & 5 & 0.946 & 0.968 & 0.967 & 0.753 & 5.2 \\
        6 & best & 10 & 0.999 & 0.992 & 0.986 & \textbf{0.869} & \textbf{5.82} \\
        \hline
    \end{tabularx}
\medskip
\label{table:results_zone_9}
\end{table}

\begin{table}[hbt!]
\caption{Results in zone 10 in terms of coverage index (for values of $\psi$ 3\%, 5\%, 10\%, and 20\%) and entropy}
\centering
    \footnotesize
    \begin{tabularx}{330pt}{cccccccc}
        \toprule
        \textbf{Line of sight} & \textbf{Initial position} & \textbf{\# patrols} &\textbf{$|W_{3}|$} & \textbf{$|W_{5}|$} & \textbf{$|W_{10}|$} & \textbf{$|W_{20}|$} & Entropy \\
        \midrule
        Greedy & random & 2 & 0.141 & 0.139 & 0.148 & 0.125 & 5.76 \\
        Greedy & random & 5 & 0.296 & 0.289 & 0.309 & 0.274 & 5.93 \\
        Greedy & random & 10 & 0.460 & 0.458 & 0.504 & 0.469 & 6.08 \\
        Greedy & best & 2 & 0.290 & 0.235 & 0.165 & 0.097 & 0.00 \\
        Greedy & best & 5 & 0.548 & 0.529 & 0.495 & 0.329 & 0.00 \\
        Greedy & best & 10 & \textbf{0.806} & \textbf{0.706} & \textbf{0.689} & 0.502 & 0.00 \\
        \midrule
        1 & random & 2 & 0.424 & 0.362 & 0.255 & 0.165 & 5.59 \\
        1 & random & 5 & 0.263 & 0.234 & 0.221 & 0.205 & 6.01 \\
        1 & random & 10 & 0.287 & 0.267 & 0.262 & 0.268 & 5.92 \\
        1 & best & 2 & 0.249 & 0.255 & 0.231 & 0.163 & 4.93 \\
        1 & best & 5 & 0.502 & 0.446 & 0.349 & 0.291 & 5.77 \\
        1 & best & 10 & 0.507 & 0.509 & 0.513 & 0.464 & 6.34 \\
        \midrule
        3 & random & 2 & 0.56 & 0.421 & 0.306 & 0.198 & 5.62 \\
        3 & random & 5 & 0.519 & 0.429 & 0.348 & 0.27 & 6.17 \\
        3 & random & 10 & 0.535 & 0.491 & 0.432 & 0.367 & 6.08 \\
        3 & best & 2 & 0.441 & 0.439 & 0.317 & 0.235 & 4.63 \\
        3 & best & 5 & 0.576 & 0.499 & 0.400 & 0.326 & 5.82 \\
        3 & best & 10 & 0.591 & 0.553 & 0.514 & 0.495 & \textbf{6.37}\\
        \midrule
        6 & random & 2 & 0.573 & 0.43 & 0.297 & 0.198 & 5.70 \\
        6 & random & 5 & 0.712 & 0.594 & 0.458 & 0.338 & 6.24 \\
        6 & random & 10 & 0.618 & 0.544 & 0.49 & 0.414 & 6.30 \\
        6 & best & 2 & 0.608 & 0.500 & 0.367 & 0.248 & 4.82 \\
        6 & best & 5 & 0.609 & 0.570 & 0.459 & 0.357 & 5.76 \\
        6 & best & 10 & 0.534 & 0.552 & 0.571 & \textbf{0.538} & 6.36\\
        \hline
    \end{tabularx}
\medskip
\label{table:results_zone_10}
\end{table}

\subsection{Results}

The model training performance is illustrated in Figures~\ref{fig:UnionGraficasArea3},~\ref{fig:UnionGraficasArea9}, and
~\ref{fig:UnionGraficasArea10}, where it can be observed that the model converges in all instances. Regarding the areas, in both zones 3 and 9, it can be observed that the configurations with 10 patrols are the least effective in terms of reward value. This occurs because the patrols are penalized for not doing enough individually when there are fewer cells to cover, as the most relevant cells are being covered by other patrols. In configurations with fewer patrols, each patrol covers more critical areas throughout the simulation, resulting in fewer penalties. For zone 10, the parameters of the reward function were changed (the coverage factor was increased from $-25$ to $-10$ and both exploration rewards were doubled) to improve the model exploration capabilities and to allow agents to identify the most relevant areas on the map. By increasing this value, the results with 10 patrols surpass those with 2 and 5 patrols.

The performance of our model in terms of the proposed metrics can be seen in Tables~\ref{table:results_zone_3},~\ref{table:results_zone_9}, and ~\ref{table:results_zone_10}, where the maximum values of each column are marked in bold, and the following $\psi$ values have been used: 3\%, 5\%, 10\%, and 20\%. The set of nodes in the $3\%$ of coverage index is the one with the highest incidence of crime; for this reason, the smaller the value of $\psi$, the more crucial achieving greater coverage becomes. Note also that as the number of nodes expected to be covered increases, the coverage decreases because of the lack of resources to cover them all, given the constraints of a connected path in the graph for each agent.

Finally, it must be pointed out that all simulations were carried out on a machine with 128 GB of RAM, 20 cores of CPU, and an RTX 4090 graphic card. Training time varies depending on the algorithm used and the specific parameter configuration for both termination and weight update of the network. In the final configurations, training times ranged between one and eight hours, for a termination condition of 100,000,000 steps. The training time varied depending on the number of patrols, the line of sight, and the size of the area, but the training time remains between three and six hours. Once trained and loaded, the model is able to generate a coordinated route in less than five minutes or provide the recommended direction in a given situation in real time.

\section{Discussion}
In the context of patrol routing problems, we have not found metrics that calculate whether the hotspots of the surveillance area have been covered or not. The most common metrics in this field to evaluate the performance are the average duration a node remains unvisited (idle time) or the average visit frequency per node. These metrics are appropriate for timed patrolling, a subcategory within the realm of patrol routing, because the objective is to repeatedly visit all the nodes of the graph or a select group of them that has been previously chosen \citep{sampaio2010gravitational}. Other metrics aim to identify the most cost-effective routes in terms of agent or fuel cost. However, once again, such problems typically have predetermined objectives to cover. Our problem does not fall within that former subcategory, and our objectives differ from those mentioned for said metrics. Firstly, our model is solved in a single shift duration, and the time spent at each node is also a relevant factor. Therefore, it would be unusual for there to be cycles within the proposed route. Furthermore, the targets to be monitored have not been preselected. Agents must dynamically determine which nodes are worth watching and which ones to exclude from their coordinated route. This differs from other proposals in which the targets, either all the nodes to be monitored or a few of them, have been predetermined in advance \citep{huang2019survey, pasqualetti2012cooperative, pasqualetti2012cooperative2, STRANDERS201363}. For all these reasons, and not having found any suitable metric, we decided to introduce the coverage index as a way of exploring the degree of coverage for the values of the target function of the graph nodes. For the police patrol problem, this metric quantifies whether agents have visited the areas with the highest crime incidence in the scenario tested. The smaller values of the percentage coverage factor, $\psi$, are the most significant, as they indicate areas with the highest concentration of crime.

 In the following, we discuss the performance results of the model in terms of the observability, number of patrols, and starting position.

\textit{Observability}. Results with a value of 1 line of sight are subject to a large variability in terms of performance. For example, in zone 9 with 10 patrols and a randomly set starting position, the model did not train. This was mainly due to the fact that agents had very limited information to coordinate with, particularly on the status of cells at a moderate distance from them, which can lead to coordination errors. This issue is particularly highlighted in large zones, such as zone 10, where the high-crime nodes are sparse. On the contrary, the performance with a greater line of sight, e.g., 3 and 6, tend to be more stable in all three areas. In zones 3 and 9, with those line-of-sight values, nearly all scenarios achieved a value over $80\%$ in the $3\%$ coverage index, $|W_{3}|$, with this value increasing as the number of patrols increased. Coverage also exceeded $70\%$ in $5\%$ of cases, with results exceeding $80\%$ when at least 5 patrols were used. The results in these two areas outperform those obtained by the greedy algorithm in all cases and for all values of coverage index tested. Nonetheless, in zone 10, the greedy algorithm performs well, and the results from our model have not been particularly noteworthy, although they continue to outperform the greedy algorithm in almost all cases. This may suggest the need to add more communication pathways for the agents, increase the number of steps, or further refine the parameters to allow better training. It is worth noting that zone 10 has about $40\%$ less crime density because it is $50\%$ larger than the other two zones and has approximately twice as many vertices and edges, which makes training the model much more difficult.

\textit{Starting position}. Regarding the starting node of the agents, the agents can be placed in the best node not yet chosen by any other agent (initial position best) or it can be random (initial position random). The purpose of this comparison is to determine whether changes in the initial positioning lead to greater diversity in the nodes visited in the routes without compromising the primary objective of crime surveillance. The results show that when random deployment is used, the increase in entropy is generally less than $10\%$ and gives worse results in configurations with 10 patrols. This can be understood by the fact that, once the optimal nodes have been initially covered, the model is not compelled to direct multiple patrols to those same locations. As a result, the randomness of the routes increases, since the optimal cell for further movement is less clearly defined (Tables~\ref{table:results_zone_3},~\ref{table:results_zone_9}, and~\ref{table:results_zone_10}). In addition, these entropy values do not outweigh the better results obtained by placing the patrols in the hotspot areas in advance. It is important to emphasize that randomness, while relevant, is not the main objective. While it is important that the routes differ from each other, the hotspots on the map remain the same, and in each iteration an attempt is made to monitor these areas over those of lesser interest to the police. This results in certain nodes being visited one or more times in each iteration of the model.

\textit{umber of patrols}. Analyzing the effects of the number of patrols, the use of only two tends to be insufficient—for the three zones, these configurations are outperformed by the configurations with 5 and 10 patrols (Tables~\ref{table:results_zone_3},~\ref{table:results_zone_9}, and~\ref{table:results_zone_10}). This seems logical, because it is not the performance of individual patrols that is measured but the overall result. A positive aspect to note is that the training with the configurations with two patrols converge more quickly than the others (Figures~\ref{fig:UnionGraficasArea3},~\ref{fig:UnionGraficasArea9}, and~\ref{fig:UnionGraficasArea10}) due to the involvement of fewer agents. Moreover, the efficiency per patrol is maximized, as relatively similar results to configurations with a larger number of resources are achieved with the minimum amount of resources. This allows us to assess the effectiveness of adding a patrol to the surveillance of a zone and, consequently, to determine the optimal number of patrols required to monitor the area effectively. For example, between configurations with 5 or 10 patrols, there is not much difference in most cases. However, on average, coverage decreases by $7\%$ with 10 patrols, when patrols are initially deployed randomly compared to the initial deployment in hotspots in the area, a situation where no improvement or worsening is observed. This suggests that the additional 5 patrols are mostly performing unnecessary tasks, resulting in a waste of public resources, and prevents over-patrolling in high-tension areas. Additionally, deploying 10 patrols in one zone, while plausible as a special allocation of resources, is neither common nor realistic for continuous deployment. This would imply having more than 100 patrols deployed at all times throughout the city.

An alternative modeling approach could involve centralizing the problem, treating patrols as resources controlled by a central agent responsible for training. However, this approach was avoided because the intention is to extend the model over time, giving agents greater autonomy and decision-making capacity. In addition, centralizing the problem may increase training time as the central agent would have to discern which of its multiple decisions is suboptimal relative to the others.

\begin{figure}[h]
\centering
\includegraphics[width=7cm,height=5cm,keepaspectratio]{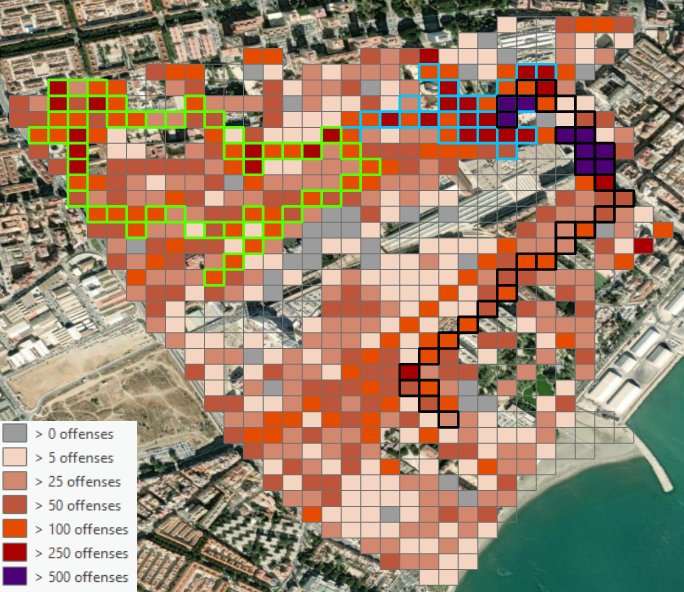}
\caption{Three of the five routes designed for our model after a simulation in zone 3.}
\label{fig:ResultsLeg}
\end{figure}

\textit{Limitations}. In terms of limitations of the proposal, it should be noted that the graphs generated from the test environments are large in size and very irregular as they represent areas of the real world. These characteristics make finding the optimal route from a given starting position prohibitively expensive, as movements are not limited to nodes already visited, there is no designated endpoint, and agents are encouraged to revisit certain nodes. Moreover, our model is not yet sufficiently developed to accommodate responses to incidents that may occur dynamically during a patrol's routine itinerary, such as responding to an accident or moving an offender to the police station. Furthermore, our environments represent medium-sized urban zones, ranging from approximately 1 to 3 km\textsuperscript{2}; these are the features of the environments that we believe are suitable for the proposed model. Figure~\ref{fig:ResultsLeg} illustrates an example of the routes designed for three patrols in zone 3, represented through different colors (lime, cyan, and black) of the cell perimeter. Also, a heat map of the entire environment has been used to visualize the crimes occurred in each cell (i.e., the value of the target function). As can be seen, the patrols manage to cover the areas with the highest crime rates. But we are aware that there are techniques more suitable for smaller map sizes, such as the timed patrolling technique \citep{pasqualetti2012cooperative, pasqualetti2012cooperative2}, primarily because, in very small areas, it may be possible to monitor all the spaces in the zone in a cyclical manner. Our proposal may not be suitable for non-urban environments where law enforcement agencies oversee several villages, farms, or forests within the surveillance area, or for urban environments with extremely sparse data, and we do not believe that the proposed approach will perform adequately in such contexts without substantial modifications to some of its key components. \citet{KADAR2019107} establishes that crimes in rural areas are largely determined by geographic points rather than other factors such as criminality data (our case) or population data. The graph resulting from that rural area would involve numerous nodes without information, which would imply a shift in the training environment from dense rewards to sparse rewards. Finally, although we believe our approach can be applied to other similar problems, in each case, it will be necessary to adjust the reward function to enable effective training for the respective problem. 

\section{Conclusions and Future Works}

This paper proposes a MARL-based model for designing patrol routing strategies, based on a decentralized partially observable Markov decision process and in terms of a target function. The model designs unpredictable and optimized routes for patrols in urban environments represented as undirected graphs, providing area coverage cooperatively, without minimal overlaps, and with partial or full observability. It generates routes suitable for medium-sized urban areas, feasible for foot patrols with a time limited working day. The model has been tested in three districts of the city of Málaga with the aim of optimizing crime surveillance. se, spatial graphs with nodes representing areas of 50 x 50 m were au×omatically generated from the street map, and we counted with rreliedaonthe crimes comed in the city, which were used as a target function of the model.

WFurthermore,e have also introduced a novel metric, the \textit{coverage index}, to evaluate the coverage performance of the routes. This index is inspired by the PAI, a well-known metric in criminology that focuses on detection of crime hotspots. Since our model does not preset routes but learns coordinated paths that cover the areas with the highest crime rates, the most commonly used performance criterion, idleness, was not appropriate for our approach. Therefore, we have performed the evaluation of the model in terms of the proposed metric. We have analyzed the impact of the varying information levels observed by agents, the number of agents, and their starting positions. Our findings show that there is no difference in performance between using five and ten patrols. This proves that such a high number of patrols is unnecessary for effective area coverage. Furthermore, results show that starting with a moderate line of sight yields results comparable to those with a larger range. This implies that agents do not need extensive information about the system to operate effectively. Random deployment has proven ineffective at generating significantly more route variety compared to deployment in optimal areas of the map, yielding poorer results on the target metric. Results suggest that the most effective approach is to deploy patrols initially at hotspots within the monitored area. A comparative study with a greedy algorithm proves that the model outperforms the greedy approach in most cases, except in one of the zones. This highlights the need to tailor parameter values to each area to ensure both convergence and task accomplishment.

In future works, we would like to explore the possibility of providing agents with a larger number of steps, thereby reducing the monitoring time at each node. This would enable agents to visit a greater number of nodes in larger urban areas. Second, we plan to compare the results obtained using neural networks with GRUs with a network architecture based on transformers, which have demonstrated success in other fields beyond their original purpose. Lastly, we will study the dynamic modification of the number of agents deployed in an area during the simulation. This scenario is quite common, as certain incidents or emergencies that could occur during surveillance may force patrols to deviate from their planned routes to attend to thefts, accidents, and other events.

\section*{Acknowledgement}
This research is partially supported by the Spanish Ministry of Science and Innovation and by the European Regional Development Fund (FEDER), Junta de Andalucía, and Universidad de Málaga through the research projects with reference PID2021-122381OB-I00 and UMA20-FEDERJA-065.
The authors want to express special thanks to the Territorial Intelligence-Analysis Group of the National Police Force (Unidad Territorial de Inteligencia del Cuerpo Nacional de Policía, UTI-CNP) and to the main head for this LEA in the province of Málaga for their support and advice for this work.



\bibliographystyle{elsarticle-harv} 
\bibliography{mybibfile}

\begin{thebibliography}{66}
\expandafter\ifx\csname natexlab\endcsname\relax\def\natexlab#1{#1}\fi
\providecommand{\url}[1]{\texttt{#1}}
\providecommand{\href}[2]{#2}
\providecommand{\path}[1]{#1}
\providecommand{\DOIprefix}{doi:}
\providecommand{\ArXivprefix}{arXiv:}
\providecommand{\URLprefix}{URL: }
\providecommand{\Pubmedprefix}{pmid:}
\providecommand{\doi}[1]{\href{http://dx.doi.org/#1}{\path{#1}}}
\providecommand{\Pubmed}[1]{\href{pmid:#1}{\path{#1}}}
\providecommand{\bibinfo}[2]{#2}
\ifx\xfnm\relax \def\xfnm[#1]{\unskip,\space#1}\fi
\bibitem[{Abbaspour et~al.(2020)Abbaspour, Fotouhi, Sedaghatbaf, Fotouhi, Vahabi and Linden}]{abbaspour2020comparative}
\bibinfo{author}{Abbaspour, S.}, \bibinfo{author}{Fotouhi, F.}, \bibinfo{author}{Sedaghatbaf, A.}, \bibinfo{author}{Fotouhi, H.}, \bibinfo{author}{Vahabi, M.}, \bibinfo{author}{Linden, M.}, \bibinfo{year}{2020}.
\newblock \bibinfo{title}{A comparative analysis of hybrid deep learning models for human activity recognition}.
\newblock \bibinfo{journal}{Sensors} \bibinfo{volume}{20}, \bibinfo{pages}{5707}.
\bibitem[{Adepeju et~al.(2016)Adepeju, Rosser and Cheng}]{adepeju2016novel}
\bibinfo{author}{Adepeju, M.}, \bibinfo{author}{Rosser, G.}, \bibinfo{author}{Cheng, T.}, \bibinfo{year}{2016}.
\newblock \bibinfo{title}{Novel evaluation metrics for sparse spatio-temporal point process hotspot predictions-a crime case study}.
\newblock \bibinfo{journal}{International Journal of Geographical Information Science} \bibinfo{volume}{30}, \bibinfo{pages}{2133--2154}.
\bibitem[{Albrecht et~al.(2024)Albrecht, Christianos and Sch{\"a}fer}]{albrecht2024multi}
\bibinfo{author}{Albrecht, S.V.}, \bibinfo{author}{Christianos, F.}, \bibinfo{author}{Sch{\"a}fer, L.}, \bibinfo{year}{2024}.
\newblock \bibinfo{title}{Multi-agent reinforcement learning: Foundations and modern approaches}.
\newblock \bibinfo{publisher}{MIT Press}.
\bibitem[{Almeida et~al.(2003)Almeida, Castro, Menezes and Ramalho}]{almeida2003combining}
\bibinfo{author}{Almeida, A.}, \bibinfo{author}{Castro, P.}, \bibinfo{author}{Menezes, T.}, \bibinfo{author}{Ramalho, G.}, \bibinfo{year}{2003}.
\newblock \bibinfo{title}{Combining idleness and distance to design heuristic agents for the patrolling task}, in: \bibinfo{booktitle}{II Brazilian Workshop in Games and Digital Entertainment}, pp. \bibinfo{pages}{33--40}.
\bibitem[{Birks et~al.(2008)Birks, Donkin and Wellsmith}]{birks2008synthesis}
\bibinfo{author}{Birks, D.J.}, \bibinfo{author}{Donkin, S.}, \bibinfo{author}{Wellsmith, M.}, \bibinfo{year}{2008}.
\newblock \bibinfo{title}{Synthesis over analysis: Towards an ontology for volume crime simulation}, in: \bibinfo{booktitle}{Artificial crime analysis systems: Using computer simulations and geographic information systems}. \bibinfo{publisher}{IGI Global}, pp. \bibinfo{pages}{160--192}.
\bibitem[{Chainey et~al.(2008)Chainey, Tompson and Uhlig}]{chainey2008utility}
\bibinfo{author}{Chainey, S.}, \bibinfo{author}{Tompson, L.}, \bibinfo{author}{Uhlig, S.}, \bibinfo{year}{2008}.
\newblock \bibinfo{title}{The utility of hotspot mapping for predicting spatial patterns of crime}.
\newblock \bibinfo{journal}{Security journal} \bibinfo{volume}{21}, \bibinfo{pages}{4--28}.
\bibitem[{Chen et~al.(2015)Chen, Cheng and Wise}]{chen2015designing}
\bibinfo{author}{Chen, H.}, \bibinfo{author}{Cheng, T.}, \bibinfo{author}{Wise, S.}, \bibinfo{year}{2015}.
\newblock \bibinfo{title}{Designing daily patrol routes for policing based on ant colony algorithm}.
\newblock \bibinfo{journal}{ISPRS Annals of the Photogrammetry, Remote Sensing and Spatial Information Sciences} \bibinfo{volume}{2}, \bibinfo{pages}{103--109}.
\bibitem[{Chen et~al.(2023)Chen, Wu, Wang, Zheng, Ma and Zhou}]{chen2023risk}
\bibinfo{author}{Chen, H.}, \bibinfo{author}{Wu, Y.}, \bibinfo{author}{Wang, W.}, \bibinfo{author}{Zheng, Z.}, \bibinfo{author}{Ma, J.}, \bibinfo{author}{Zhou, B.}, \bibinfo{year}{2023}.
\newblock \bibinfo{title}{A risk-aware multi-objective patrolling route optimization method using reinforcement learning}, in: \bibinfo{booktitle}{2023 IEEE 29th International Conference on Parallel and Distributed Systems (ICPADS)}, \bibinfo{organization}{IEEE}. pp. \bibinfo{pages}{1637--1644}.
\bibitem[{Chevaleyre(2004)}]{chevaleyre2004theoretical}
\bibinfo{author}{Chevaleyre, Y.}, \bibinfo{year}{2004}.
\newblock \bibinfo{title}{Theoretical analysis of the multi-agent patrolling problem}, in: \bibinfo{booktitle}{Proceedings. IEEE/WIC/ACM International Conference on Intelligent Agent Technology, 2004.(IAT 2004).}, \bibinfo{organization}{IEEE}. pp. \bibinfo{pages}{302--308}.
\bibitem[{Cho et~al.(2014)Cho, Van~Merri{\"e}nboer, Gulcehre, Bahdanau, Bougares, Schwenk and Bengio}]{cho2014learning}
\bibinfo{author}{Cho, K.}, \bibinfo{author}{Van~Merri{\"e}nboer, B.}, \bibinfo{author}{Gulcehre, C.}, \bibinfo{author}{Bahdanau, D.}, \bibinfo{author}{Bougares, F.}, \bibinfo{author}{Schwenk, H.}, \bibinfo{author}{Bengio, Y.}, \bibinfo{year}{2014}.
\newblock \bibinfo{title}{Learning phrase representations using rnn encoder-decoder for statistical machine translation}.
\newblock \bibinfo{journal}{arXiv preprint arXiv:1406.1078} .
\bibitem[{De~Witt et~al.(2020)De~Witt, Gupta, Makoviichuk, Makoviychuk, Torr, Sun and Whiteson}]{de2020independent}
\bibinfo{author}{De~Witt, C.S.}, \bibinfo{author}{Gupta, T.}, \bibinfo{author}{Makoviichuk, D.}, \bibinfo{author}{Makoviychuk, V.}, \bibinfo{author}{Torr, P.H.}, \bibinfo{author}{Sun, M.}, \bibinfo{author}{Whiteson, S.}, \bibinfo{year}{2020}.
\newblock \bibinfo{title}{Is independent learning all you need in the starcraft multi-agent challenge?}
\newblock \bibinfo{journal}{arXiv preprint arXiv:2011.09533} .
\bibitem[{Devia and Weber(2013)}]{devia2013generating}
\bibinfo{author}{Devia, N.}, \bibinfo{author}{Weber, R.}, \bibinfo{year}{2013}.
\newblock \bibinfo{title}{Generating crime data using agent-based simulation}.
\newblock \bibinfo{journal}{Computers, Environment and Urban Systems} \bibinfo{volume}{42}, \bibinfo{pages}{26--41}.
\bibitem[{Eck(1993)}]{eck1993threat}
\bibinfo{author}{Eck, J.E.}, \bibinfo{year}{1993}.
\newblock \bibinfo{title}{The threat of crime displacement}, in: \bibinfo{booktitle}{Criminal justice abstracts}, pp. \bibinfo{pages}{527--546}.
\bibitem[{Guo et~al.(2023)Guo, Pan, Duan and He}]{guo2023balancing}
\bibinfo{author}{Guo, L.}, \bibinfo{author}{Pan, H.}, \bibinfo{author}{Duan, X.}, \bibinfo{author}{He, J.}, \bibinfo{year}{2023}.
\newblock \bibinfo{title}{Balancing efficiency and unpredictability in multi-robot patrolling: A marl-based approach}, in: \bibinfo{booktitle}{2023 IEEE International Conference on Robotics and Automation (ICRA)}, \bibinfo{organization}{IEEE}. pp. \bibinfo{pages}{3504--3509}.
\bibitem[{Hari et~al.(2020)Hari, Rathinam, Darbha, Kalyanam, Manyam and Casbeer}]{hari2020optimal}
\bibinfo{author}{Hari, S.K.K.}, \bibinfo{author}{Rathinam, S.}, \bibinfo{author}{Darbha, S.}, \bibinfo{author}{Kalyanam, K.}, \bibinfo{author}{Manyam, S.G.}, \bibinfo{author}{Casbeer, D.}, \bibinfo{year}{2020}.
\newblock \bibinfo{title}{Optimal uav route planning for persistent monitoring missions}.
\newblock \bibinfo{journal}{IEEE Transactions on Robotics} \bibinfo{volume}{37}, \bibinfo{pages}{550--566}.
\bibitem[{Hochreiter and Schmidhuber(1997)}]{hochreiter1997long}
\bibinfo{author}{Hochreiter, S.}, \bibinfo{author}{Schmidhuber, J.}, \bibinfo{year}{1997}.
\newblock \bibinfo{title}{Long short-term memory}.
\newblock \bibinfo{journal}{Neural computation} \bibinfo{volume}{9}, \bibinfo{pages}{1735--1780}.
\bibitem[{Hu et~al.(2023)Hu, Zhong, Gao, Wang, Dong, Liang, Li, Chang and Yang}]{hu2022marllib}
\bibinfo{author}{Hu, S.}, \bibinfo{author}{Zhong, Y.}, \bibinfo{author}{Gao, M.}, \bibinfo{author}{Wang, W.}, \bibinfo{author}{Dong, H.}, \bibinfo{author}{Liang, X.}, \bibinfo{author}{Li, Z.}, \bibinfo{author}{Chang, X.}, \bibinfo{author}{Yang, Y.}, \bibinfo{year}{2023}.
\newblock \bibinfo{title}{Marllib: A scalable and efficient multi-agent reinforcement learning library}.
\newblock \bibinfo{journal}{Journal of Machine Learning Research} .
\bibitem[{Huang et~al.(2019)Huang, Zhou, Hao and Hou}]{huang2019survey}
\bibinfo{author}{Huang, L.}, \bibinfo{author}{Zhou, M.}, \bibinfo{author}{Hao, K.}, \bibinfo{author}{Hou, E.}, \bibinfo{year}{2019}.
\newblock \bibinfo{title}{A survey of multi-robot regular and adversarial patrolling}.
\newblock \bibinfo{journal}{IEEE/CAA Journal of Automatica Sinica} \bibinfo{volume}{6}, \bibinfo{pages}{894--903}.
\bibitem[{Hwang et~al.(2009)Hwang, Lin and Huang}]{hwang2009cooperative}
\bibinfo{author}{Hwang, K.S.}, \bibinfo{author}{Lin, J.L.}, \bibinfo{author}{Huang, H.L.}, \bibinfo{year}{2009}.
\newblock \bibinfo{title}{Cooperative patrol planning of multi-robot systems by a competitive auction system}, in: \bibinfo{booktitle}{2009 ICCAS-SICE}, \bibinfo{organization}{IEEE}. pp. \bibinfo{pages}{4359--4363}.
\bibitem[{Iqbal and Sha(2019)}]{iqbal2019actor}
\bibinfo{author}{Iqbal, S.}, \bibinfo{author}{Sha, F.}, \bibinfo{year}{2019}.
\newblock \bibinfo{title}{Actor-attention-critic for multi-agent reinforcement learning}, in: \bibinfo{booktitle}{International conference on machine learning}, \bibinfo{organization}{PMLR}. pp. \bibinfo{pages}{2961--2970}.
\bibitem[{Kadar et~al.(2019)Kadar, Maculan and Feuerriegel}]{KADAR2019107}
\bibinfo{author}{Kadar, C.}, \bibinfo{author}{Maculan, R.}, \bibinfo{author}{Feuerriegel, S.}, \bibinfo{year}{2019}.
\newblock \bibinfo{title}{Public decision support for low population density areas: An imbalance-aware hyper-ensemble for spatio-temporal crime prediction}.
\newblock \bibinfo{journal}{Decision Support Systems} \bibinfo{volume}{119}, \bibinfo{pages}{107--117}.
\newblock \DOIprefix\doi{https://doi.org/10.1016/j.dss.2019.03.001}.
\bibitem[{Kuba et~al.(2021)Kuba, Chen, Wen, Wen, Sun, Wang and Yang}]{kuba2021trust}
\bibinfo{author}{Kuba, J.G.}, \bibinfo{author}{Chen, R.}, \bibinfo{author}{Wen, M.}, \bibinfo{author}{Wen, Y.}, \bibinfo{author}{Sun, F.}, \bibinfo{author}{Wang, J.}, \bibinfo{author}{Yang, Y.}, \bibinfo{year}{2021}.
\newblock \bibinfo{title}{Trust region policy optimisation in multi-agent reinforcement learning}.
\newblock \bibinfo{journal}{arXiv preprint arXiv:2109.11251} .
\bibitem[{Lee and Lee(2021)}]{LEE2021296}
\bibinfo{author}{Lee, H.R.}, \bibinfo{author}{Lee, T.}, \bibinfo{year}{2021}.
\newblock \bibinfo{title}{Multi-agent reinforcement learning algorithm to solve a partially-observable multi-agent problem in disaster response}.
\newblock \bibinfo{journal}{European Journal of Operational Research} \bibinfo{volume}{291}, \bibinfo{pages}{296--308}.
\newblock \DOIprefix\doi{https://doi.org/10.1016/j.ejor.2020.09.018}.
\bibitem[{Lee et~al.(2020)Lee, SooHyun and Eck}]{lee2020theory}
\bibinfo{author}{Lee, Y.}, \bibinfo{author}{SooHyun, O.}, \bibinfo{author}{Eck, J.E.}, \bibinfo{year}{2020}.
\newblock \bibinfo{title}{A theory-driven algorithm for real-time crime hot spot forecasting}.
\newblock \bibinfo{journal}{Police Quarterly} \bibinfo{volume}{23}, \bibinfo{pages}{174--201}.
\bibitem[{Li and He(2023)}]{li2023multiagent}
\bibinfo{author}{Li, H.}, \bibinfo{author}{He, H.}, \bibinfo{year}{2023}.
\newblock \bibinfo{title}{Multiagent trust region policy optimization}.
\newblock \bibinfo{journal}{IEEE Transactions on Neural Networks and Learning Systems} .
\bibitem[{Liang et~al.(2018)Liang, Liaw, Nishihara, Moritz, Fox, Goldberg, Gonzalez, Jordan and Stoica}]{liang2018rllib}
\bibinfo{author}{Liang, E.}, \bibinfo{author}{Liaw, R.}, \bibinfo{author}{Nishihara, R.}, \bibinfo{author}{Moritz, P.}, \bibinfo{author}{Fox, R.}, \bibinfo{author}{Goldberg, K.}, \bibinfo{author}{Gonzalez, J.E.}, \bibinfo{author}{Jordan, M.I.}, \bibinfo{author}{Stoica, I.}, \bibinfo{year}{2018}.
\newblock \bibinfo{title}{{RLlib}: Abstractions for distributed reinforcement learning}, in: \bibinfo{booktitle}{International Conference on Machine Learning ({ICML})}.
\bibitem[{Lillicrap et~al.(2015)Lillicrap, Hunt, Pritzel, Heess, Erez, Tassa, Silver and Wierstra}]{lillicrap2015continuous}
\bibinfo{author}{Lillicrap, T.P.}, \bibinfo{author}{Hunt, J.J.}, \bibinfo{author}{Pritzel, A.}, \bibinfo{author}{Heess, N.}, \bibinfo{author}{Erez, T.}, \bibinfo{author}{Tassa, Y.}, \bibinfo{author}{Silver, D.}, \bibinfo{author}{Wierstra, D.}, \bibinfo{year}{2015}.
\newblock \bibinfo{title}{Continuous control with deep reinforcement learning}.
\newblock \bibinfo{journal}{arXiv preprint arXiv:1509.02971} .
\bibitem[{Lowe et~al.(2017)Lowe, Wu, Tamar, Harb, Pieter~Abbeel and Mordatch}]{lowe2017multi}
\bibinfo{author}{Lowe, R.}, \bibinfo{author}{Wu, Y.I.}, \bibinfo{author}{Tamar, A.}, \bibinfo{author}{Harb, J.}, \bibinfo{author}{Pieter~Abbeel, O.}, \bibinfo{author}{Mordatch, I.}, \bibinfo{year}{2017}.
\newblock \bibinfo{title}{Multi-agent actor-critic for mixed cooperative-competitive environments}.
\newblock \bibinfo{journal}{Advances in neural information processing systems} \bibinfo{volume}{30}.
\bibitem[{Luis et~al.(2022)Luis, Peralta, C{\'o}rdoba, del Nozal, Mar{\'\i}n and Reina}]{luis2022evolutionary}
\bibinfo{author}{Luis, S.Y.}, \bibinfo{author}{Peralta, F.}, \bibinfo{author}{C{\'o}rdoba, A.T.}, \bibinfo{author}{del Nozal, {\'A}.R.}, \bibinfo{author}{Mar{\'\i}n, S.T.}, \bibinfo{author}{Reina, D.G.}, \bibinfo{year}{2022}.
\newblock \bibinfo{title}{An evolutionary multi-objective path planning of a fleet of asvs for patrolling water resources}.
\newblock \bibinfo{journal}{Engineering Applications of Artificial Intelligence} \bibinfo{volume}{112}, \bibinfo{pages}{104852}.
\bibitem[{Ma and Luo(2022)}]{ma2022value}
\bibinfo{author}{Ma, Y.}, \bibinfo{author}{Luo, J.}, \bibinfo{year}{2022}.
\newblock \bibinfo{title}{Value-decomposition multi-agent proximal policy optimization}, in: \bibinfo{booktitle}{2022 China Automation Congress (CAC)}, \bibinfo{organization}{IEEE}. pp. \bibinfo{pages}{3460--3464}.
\bibitem[{Machado et~al.(2002a)Machado, Almeida, Ramalho, Zucker and Drogoul}]{machado2002multicoor}
\bibinfo{author}{Machado, A.}, \bibinfo{author}{Almeida, A.}, \bibinfo{author}{Ramalho, G.}, \bibinfo{author}{Zucker, J.D.}, \bibinfo{author}{Drogoul, A.}, \bibinfo{year}{2002}a.
\newblock \bibinfo{title}{Multi-agent movement coordination in patrolling}, in: \bibinfo{booktitle}{Proceedings of the 3rd International Conference on Computer and Game}, pp. \bibinfo{pages}{155--170}.
\bibitem[{Machado et~al.(2002b)Machado, Ramalho, Zucker and Drogoul}]{machado2002multi}
\bibinfo{author}{Machado, A.}, \bibinfo{author}{Ramalho, G.}, \bibinfo{author}{Zucker, J.D.}, \bibinfo{author}{Drogoul, A.}, \bibinfo{year}{2002}b.
\newblock \bibinfo{title}{Multi-agent patrolling: An empirical analysis of alternative architectures}, in: \bibinfo{booktitle}{International workshop on multi-agent systems and agent-based simulation}, \bibinfo{organization}{Springer}. pp. \bibinfo{pages}{155--170}.
\bibitem[{Mnih et~al.(2016)Mnih, Badia, Mirza, Graves, Lillicrap, Harley, Silver and Kavukcuoglu}]{mnih2016asynchronous}
\bibinfo{author}{Mnih, V.}, \bibinfo{author}{Badia, A.P.}, \bibinfo{author}{Mirza, M.}, \bibinfo{author}{Graves, A.}, \bibinfo{author}{Lillicrap, T.}, \bibinfo{author}{Harley, T.}, \bibinfo{author}{Silver, D.}, \bibinfo{author}{Kavukcuoglu, K.}, \bibinfo{year}{2016}.
\newblock \bibinfo{title}{Asynchronous methods for deep reinforcement learning}, in: \bibinfo{booktitle}{International conference on machine learning}, \bibinfo{organization}{PMLR}. pp. \bibinfo{pages}{1928--1937}.
\bibitem[{Mnih et~al.(2015)Mnih, Kavukcuoglu, Silver, Rusu, Veness, Bellemare, Graves, Riedmiller, Fidjeland, Ostrovski et~al.}]{mnih2015human}
\bibinfo{author}{Mnih, V.}, \bibinfo{author}{Kavukcuoglu, K.}, \bibinfo{author}{Silver, D.}, \bibinfo{author}{Rusu, A.A.}, \bibinfo{author}{Veness, J.}, \bibinfo{author}{Bellemare, M.G.}, \bibinfo{author}{Graves, A.}, \bibinfo{author}{Riedmiller, M.}, \bibinfo{author}{Fidjeland, A.K.}, \bibinfo{author}{Ostrovski, G.}, et~al., \bibinfo{year}{2015}.
\newblock \bibinfo{title}{Human-level control through deep reinforcement learning}.
\newblock \bibinfo{journal}{nature} \bibinfo{volume}{518}, \bibinfo{pages}{529--533}.
\bibitem[{{Málaga city council}(2024)}]{malagaSistemaInformacion}
\bibinfo{author}{{Málaga city council}}, \bibinfo{year}{2024}.
\newblock \bibinfo{title}{{S}istema de {I}nformación {C}artográfica - {C}allejero - {D}atos abiertos {A}yto. {M}álaga --- datosabiertos.malaga.eu}.
\newblock \bibinfo{note}{[Accessed 08-04-2024]}.
\bibitem[{Oliehoek et~al.(2016)Oliehoek, Amato et~al.}]{oliehoek2016concise}
\bibinfo{author}{Oliehoek, F.A.}, \bibinfo{author}{Amato, C.}, et~al., \bibinfo{year}{2016}.
\newblock \bibinfo{title}{A concise introduction to decentralized POMDPs}. volume~\bibinfo{volume}{1}.
\newblock \bibinfo{publisher}{Springer}.
\bibitem[{{OpenStreetMap contributors}(2017)}]{OpenStreetMap}
\bibinfo{author}{{OpenStreetMap contributors}}, \bibinfo{year}{2017}.
\newblock \bibinfo{title}{{Planet dump retrieved from https://planet.osm.org }}.
\newblock \bibinfo{howpublished}{\url{ https://www.openstreetmap.org }}.
\bibitem[{Othmani-Guibourg et~al.(2018)Othmani-Guibourg, El~Fallah-Seghrouchni and Farges}]{othmani2018}
\bibinfo{author}{Othmani-Guibourg, M.}, \bibinfo{author}{El~Fallah-Seghrouchni, A.}, \bibinfo{author}{Farges, J.L.}, \bibinfo{year}{2018}.
\newblock \bibinfo{title}{Path generation with lstm recurrent neural networks in the context of the multi-agent patrolling}, in: \bibinfo{booktitle}{2018 IEEE 30th International Conference on Tools with Artificial Intelligence (ICTAI)}, pp. \bibinfo{pages}{430--437}.
\newblock \DOIprefix\doi{10.1109/ICTAI.2018.00073}.
\bibitem[{Othmani-Guibourg et~al.(2017)Othmani-Guibourg, El~Fallah-Seghrouchni, Farges and Potop-Butucaru}]{othmani2017multi}
\bibinfo{author}{Othmani-Guibourg, M.}, \bibinfo{author}{El~Fallah-Seghrouchni, A.}, \bibinfo{author}{Farges, J.L.}, \bibinfo{author}{Potop-Butucaru, M.}, \bibinfo{year}{2017}.
\newblock \bibinfo{title}{Multi-agent patrolling in dynamic environments}, in: \bibinfo{booktitle}{2017 IEEE international conference on agents (ICA)}, \bibinfo{organization}{IEEE}. pp. \bibinfo{pages}{72--77}.
\bibitem[{Othmani-Guibourg et~al.(2019)Othmani-Guibourg, Farges and Seghrouchni}]{othmani2019}
\bibinfo{author}{Othmani-Guibourg, M.}, \bibinfo{author}{Farges, J.L.}, \bibinfo{author}{Seghrouchni, A.}, \bibinfo{year}{2019}.
\newblock \bibinfo{title}{Lstm path-maker: a new lstm-based strategy for the multi-agent patrolling}, in: \bibinfo{booktitle}{HAWAII INTERNATIONAL CONFERENCE ON SYSTEM SCIENCES 2019 (HICSS-52)}.
\newblock \DOIprefix\doi{10.24251/HICSS.2019.076}.
\bibitem[{Parker et~al.(2016)Parker, Nunes, Godoy and Gini}]{parker2016exploiting}
\bibinfo{author}{Parker, J.}, \bibinfo{author}{Nunes, E.}, \bibinfo{author}{Godoy, J.}, \bibinfo{author}{Gini, M.}, \bibinfo{year}{2016}.
\newblock \bibinfo{title}{Exploiting spatial locality and heterogeneity of agents for search and rescue teamwork}.
\newblock \bibinfo{journal}{Journal of Field Robotics} \bibinfo{volume}{33}, \bibinfo{pages}{877--900}.
\bibitem[{Pasqualetti et~al.(2012a)Pasqualetti, Durham and Bullo}]{pasqualetti2012cooperative}
\bibinfo{author}{Pasqualetti, F.}, \bibinfo{author}{Durham, J.W.}, \bibinfo{author}{Bullo, F.}, \bibinfo{year}{2012}a.
\newblock \bibinfo{title}{Cooperative patrolling via weighted tours: Performance analysis and distributed algorithms}.
\newblock \bibinfo{journal}{IEEE Transactions on Robotics} \bibinfo{volume}{28}, \bibinfo{pages}{1181--1188}.
\bibitem[{Pasqualetti et~al.(2012b)Pasqualetti, Franchi and Bullo}]{pasqualetti2012cooperative2}
\bibinfo{author}{Pasqualetti, F.}, \bibinfo{author}{Franchi, A.}, \bibinfo{author}{Bullo, F.}, \bibinfo{year}{2012}b.
\newblock \bibinfo{title}{On cooperative patrolling: Optimal trajectories, complexity analysis, and approximation algorithms}.
\newblock \bibinfo{journal}{IEEE Transactions on Robotics} \bibinfo{volume}{28}, \bibinfo{pages}{592--606}.
\bibitem[{Peng et~al.(2021)Peng, Rashid, Schroeder~de Witt, Kamienny, Torr, B{\"o}hmer and Whiteson}]{peng2021facmac}
\bibinfo{author}{Peng, B.}, \bibinfo{author}{Rashid, T.}, \bibinfo{author}{Schroeder~de Witt, C.}, \bibinfo{author}{Kamienny, P.A.}, \bibinfo{author}{Torr, P.}, \bibinfo{author}{B{\"o}hmer, W.}, \bibinfo{author}{Whiteson, S.}, \bibinfo{year}{2021}.
\newblock \bibinfo{title}{Facmac: Factored multi-agent centralised policy gradients}.
\newblock \bibinfo{journal}{Advances in Neural Information Processing Systems} \bibinfo{volume}{34}, \bibinfo{pages}{12208--12221}.
\bibitem[{Portugal et~al.(2013)Portugal, Couceiro and Rocha}]{portugal2013applying}
\bibinfo{author}{Portugal, D.}, \bibinfo{author}{Couceiro, M.S.}, \bibinfo{author}{Rocha, R.P.}, \bibinfo{year}{2013}.
\newblock \bibinfo{title}{Applying bayesian learning to multi-robot patrol}, in: \bibinfo{booktitle}{2013 IEEE International Symposium on Safety, Security, and Rescue Robotics (SSRR)}, \bibinfo{organization}{IEEE}. pp. \bibinfo{pages}{1--6}.
\bibitem[{Portugal and Rocha(2010)}]{portugal2010msp}
\bibinfo{author}{Portugal, D.}, \bibinfo{author}{Rocha, R.}, \bibinfo{year}{2010}.
\newblock \bibinfo{title}{Msp algorithm: multi-robot patrolling based on territory allocation using balanced graph partitioning}, in: \bibinfo{booktitle}{Proceedings of the 2010 ACM symposium on applied computing}, pp. \bibinfo{pages}{1271--1276}.
\bibitem[{Portugal and Rocha(2016)}]{portugal2016cooperative}
\bibinfo{author}{Portugal, D.}, \bibinfo{author}{Rocha, R.P.}, \bibinfo{year}{2016}.
\newblock \bibinfo{title}{Cooperative multi-robot patrol with bayesian learning}.
\newblock \bibinfo{journal}{Autonomous Robots} \bibinfo{volume}{40}, \bibinfo{pages}{929--953}.
\bibitem[{Rummens et~al.(2017)Rummens, Hardyns and Pauwels}]{RUMMENS2017255}
\bibinfo{author}{Rummens, A.}, \bibinfo{author}{Hardyns, W.}, \bibinfo{author}{Pauwels, L.}, \bibinfo{year}{2017}.
\newblock \bibinfo{title}{The use of predictive analysis in spatiotemporal crime forecasting: Building and testing a model in an urban context}.
\newblock \bibinfo{journal}{Applied Geography} \bibinfo{volume}{86}, \bibinfo{pages}{255--261}.
\newblock \DOIprefix\doi{https://doi.org/10.1016/j.apgeog.2017.06.011}.
\bibitem[{Samanta et~al.(2022)Samanta, Sen and Ghosh}]{samanta2022literature}
\bibinfo{author}{Samanta, S.}, \bibinfo{author}{Sen, G.}, \bibinfo{author}{Ghosh, S.K.}, \bibinfo{year}{2022}.
\newblock \bibinfo{title}{A literature review on police patrolling problems}.
\newblock \bibinfo{journal}{Annals of Operations Research} \bibinfo{volume}{316}, \bibinfo{pages}{1063--1106}.
\bibitem[{Sampaio et~al.(2010)Sampaio, Ramalho and Tedesco}]{sampaio2010gravitational}
\bibinfo{author}{Sampaio, P.A.}, \bibinfo{author}{Ramalho, G.}, \bibinfo{author}{Tedesco, P.}, \bibinfo{year}{2010}.
\newblock \bibinfo{title}{The gravitational strategy for the timed patrolling}, in: \bibinfo{booktitle}{2010 22nd IEEE International Conference on Tools with Artificial Intelligence}, \bibinfo{organization}{IEEE}. pp. \bibinfo{pages}{113--120}.
\bibitem[{Samvelyan et~al.(2019)Samvelyan, Rashid, De~Witt, Farquhar, Nardelli, Rudner, Hung, Torr, Foerster and Whiteson}]{samvelyan2019starcraft}
\bibinfo{author}{Samvelyan, M.}, \bibinfo{author}{Rashid, T.}, \bibinfo{author}{De~Witt, C.S.}, \bibinfo{author}{Farquhar, G.}, \bibinfo{author}{Nardelli, N.}, \bibinfo{author}{Rudner, T.G.}, \bibinfo{author}{Hung, C.M.}, \bibinfo{author}{Torr, P.H.}, \bibinfo{author}{Foerster, J.}, \bibinfo{author}{Whiteson, S.}, \bibinfo{year}{2019}.
\newblock \bibinfo{title}{The starcraft multi-agent challenge}.
\newblock \bibinfo{journal}{arXiv preprint arXiv:1902.04043} .
\bibitem[{Santana et~al.(2004)Santana, Ramalho, Corruble and Ratitch}]{santana2004multi}
\bibinfo{author}{Santana, H.}, \bibinfo{author}{Ramalho, G.}, \bibinfo{author}{Corruble, V.}, \bibinfo{author}{Ratitch, B.}, \bibinfo{year}{2004}.
\newblock \bibinfo{title}{Multi-agent patrolling with reinforcement learning}, in: \bibinfo{booktitle}{Autonomous Agents and Multiagent Systems, International Joint Conference on}, \bibinfo{organization}{IEEE Computer Society}. pp. \bibinfo{pages}{1122--1129}.
\bibitem[{Schulman et~al.(2015a)Schulman, Levine, Abbeel, Jordan and Moritz}]{schulman2015trust}
\bibinfo{author}{Schulman, J.}, \bibinfo{author}{Levine, S.}, \bibinfo{author}{Abbeel, P.}, \bibinfo{author}{Jordan, M.}, \bibinfo{author}{Moritz, P.}, \bibinfo{year}{2015}a.
\newblock \bibinfo{title}{Trust region policy optimization}, in: \bibinfo{booktitle}{International conference on machine learning}, \bibinfo{organization}{PMLR}. pp. \bibinfo{pages}{1889--1897}.
\bibitem[{Schulman et~al.(2015b)Schulman, Moritz, Levine, Jordan and Abbeel}]{schulman2015high}
\bibinfo{author}{Schulman, J.}, \bibinfo{author}{Moritz, P.}, \bibinfo{author}{Levine, S.}, \bibinfo{author}{Jordan, M.}, \bibinfo{author}{Abbeel, P.}, \bibinfo{year}{2015}b.
\newblock \bibinfo{title}{High-dimensional continuous control using generalized advantage estimation}.
\newblock \bibinfo{journal}{arXiv preprint arXiv:1506.02438} .
\bibitem[{Schulman et~al.(2017)Schulman, Wolski, Dhariwal, Radford and Klimov}]{schulman2017proximal}
\bibinfo{author}{Schulman, J.}, \bibinfo{author}{Wolski, F.}, \bibinfo{author}{Dhariwal, P.}, \bibinfo{author}{Radford, A.}, \bibinfo{author}{Klimov, O.}, \bibinfo{year}{2017}.
\newblock \bibinfo{title}{Proximal policy optimization algorithms}.
\newblock \bibinfo{journal}{arXiv preprint arXiv:1707.06347} .
\bibitem[{Sea et~al.(2018)Sea, Sugiyama and Sugawara}]{sea2018frequency}
\bibinfo{author}{Sea, V.}, \bibinfo{author}{Sugiyama, A.}, \bibinfo{author}{Sugawara, T.}, \bibinfo{year}{2018}.
\newblock \bibinfo{title}{Frequency-based multi-agent patrolling model and its area partitioning solution method for balanced workload}, in: \bibinfo{booktitle}{Integration of Constraint Programming, Artificial Intelligence, and Operations Research: 15th International Conference, CPAIOR 2018, Delft, The Netherlands, June 26--29, 2018, Proceedings 15}, \bibinfo{organization}{Springer}. pp. \bibinfo{pages}{530--545}.
\bibitem[{Shalev-Shwartz et~al.(2016)Shalev-Shwartz, Shammah and Shashua}]{shalev2016safe}
\bibinfo{author}{Shalev-Shwartz, S.}, \bibinfo{author}{Shammah, S.}, \bibinfo{author}{Shashua, A.}, \bibinfo{year}{2016}.
\newblock \bibinfo{title}{Safe, multi-agent, reinforcement learning for autonomous driving}.
\newblock \bibinfo{journal}{arXiv preprint arXiv:1610.03295} .
\bibitem[{Sherman et~al.(2014)Sherman, Williams, Ariel, Strang, Wain, Slothower and Norton}]{sherman2014integrated}
\bibinfo{author}{Sherman, L.W.}, \bibinfo{author}{Williams, S.}, \bibinfo{author}{Ariel, B.}, \bibinfo{author}{Strang, L.R.}, \bibinfo{author}{Wain, N.}, \bibinfo{author}{Slothower, M.}, \bibinfo{author}{Norton, A.}, \bibinfo{year}{2014}.
\newblock \bibinfo{title}{An integrated theory of hot spots patrol strategy: implementing prevention by scaling up and feeding back}.
\newblock \bibinfo{journal}{Journal of Contemporary Criminal Justice} \bibinfo{volume}{30}, \bibinfo{pages}{95--122}.
\bibitem[{Soliman et~al.(2023)Soliman, Al-Ali, Mohamed, Gedawy, Izham, Bahri, Erbad and Guizani}]{soliman2023ai}
\bibinfo{author}{Soliman, A.}, \bibinfo{author}{Al-Ali, A.}, \bibinfo{author}{Mohamed, A.}, \bibinfo{author}{Gedawy, H.}, \bibinfo{author}{Izham, D.}, \bibinfo{author}{Bahri, M.}, \bibinfo{author}{Erbad, A.}, \bibinfo{author}{Guizani, M.}, \bibinfo{year}{2023}.
\newblock \bibinfo{title}{Ai-based uav navigation framework with digital twin technology for mobile target visitation}.
\newblock \bibinfo{journal}{Engineering Applications of Artificial Intelligence} \bibinfo{volume}{123}, \bibinfo{pages}{106318}.
\bibitem[{Stranders et~al.(2013)Stranders, {Munoz de Cote}, Rogers and Jennings}]{STRANDERS201363}
\bibinfo{author}{Stranders, R.}, \bibinfo{author}{{Munoz de Cote}, E.}, \bibinfo{author}{Rogers, A.}, \bibinfo{author}{Jennings, N.}, \bibinfo{year}{2013}.
\newblock \bibinfo{title}{Near-optimal continuous patrolling with teams of mobile information gathering agents}.
\newblock \bibinfo{journal}{Artificial Intelligence} \bibinfo{volume}{195}, \bibinfo{pages}{63--105}.
\newblock \DOIprefix\doi{https://doi.org/10.1016/j.artint.2012.10.006}.
\bibitem[{Su et~al.(2021)Su, Adams and Beling}]{su2021value}
\bibinfo{author}{Su, J.}, \bibinfo{author}{Adams, S.}, \bibinfo{author}{Beling, P.}, \bibinfo{year}{2021}.
\newblock \bibinfo{title}{Value-decomposition multi-agent actor-critics}, in: \bibinfo{booktitle}{Proceedings of the AAAI conference on artificial intelligence}, pp. \bibinfo{pages}{11352--11360}.
\bibitem[{Wijaya and Maulidevi(2019)}]{wijaya2019multiagent}
\bibinfo{author}{Wijaya, R.I.}, \bibinfo{author}{Maulidevi, N.U.}, \bibinfo{year}{2019}.
\newblock \bibinfo{title}{Multiagent system development for cooperative multiplayer video game using deep q-network}, in: \bibinfo{booktitle}{2019 International Conference of Advanced Informatics: Concepts, Theory and Applications (ICAICTA)}, \bibinfo{organization}{IEEE}. pp. \bibinfo{pages}{1--5}.
\bibitem[{Yan and Zhang(2016)}]{yan2016multi}
\bibinfo{author}{Yan, C.}, \bibinfo{author}{Zhang, T.}, \bibinfo{year}{2016}.
\newblock \bibinfo{title}{Multi-robot patrol: A distributed algorithm based on expected idleness}.
\newblock \bibinfo{journal}{International Journal of Advanced Robotic Systems} \bibinfo{volume}{13}, \bibinfo{pages}{1729881416663666}.
\bibitem[{Yin et~al.(2012)Yin, Jiang, Tambe, Kiekintveld, Leyton-Brown, Sandholm and Sullivan}]{yin2012trusts}
\bibinfo{author}{Yin, Z.}, \bibinfo{author}{Jiang, A.X.}, \bibinfo{author}{Tambe, M.}, \bibinfo{author}{Kiekintveld, C.}, \bibinfo{author}{Leyton-Brown, K.}, \bibinfo{author}{Sandholm, T.}, \bibinfo{author}{Sullivan, J.P.}, \bibinfo{year}{2012}.
\newblock \bibinfo{title}{Trusts: Scheduling randomized patrols for fare inspection in transit systems using game theory}.
\newblock \bibinfo{journal}{AI magazine} \bibinfo{volume}{33}, \bibinfo{pages}{59--59}.
\bibitem[{Yu et~al.(2022)Yu, Velu, Vinitsky, Gao, Wang, Bayen and Wu}]{yu2022surprising}
\bibinfo{author}{Yu, C.}, \bibinfo{author}{Velu, A.}, \bibinfo{author}{Vinitsky, E.}, \bibinfo{author}{Gao, J.}, \bibinfo{author}{Wang, Y.}, \bibinfo{author}{Bayen, A.}, \bibinfo{author}{Wu, Y.}, \bibinfo{year}{2022}.
\newblock \bibinfo{title}{The surprising effectiveness of ppo in cooperative multi-agent games}.
\newblock \bibinfo{journal}{Advances in Neural Information Processing Systems} \bibinfo{volume}{35}, \bibinfo{pages}{24611--24624}.
\bibitem[{Zhang et~al.(2021)Zhang, Lu, Garg and Foerster}]{zhang2021centralized}
\bibinfo{author}{Zhang, Q.}, \bibinfo{author}{Lu, C.}, \bibinfo{author}{Garg, A.}, \bibinfo{author}{Foerster, J.}, \bibinfo{year}{2021}.
\newblock \bibinfo{title}{Centralized model and exploration policy for multi-agent rl}.
\newblock \bibinfo{journal}{arXiv preprint arXiv:2107.06434} .

\end{thebibliography}

\end{document}